\documentclass[10pt]{article} % For LaTeX2e
\usepackage[preprint]{tmlr}

\usepackage{amsmath,amsfonts,bm}

\def\eqref#1{equation~\ref{#1}}
\def\1{\bm{1}}

\DeclareMathAlphabet{\mathsfit}{\encodingdefault}{\sfdefault}{m}{sl}
\SetMathAlphabet{\mathsfit}{bold}{\encodingdefault}{\sfdefault}{bx}{n}

\newcommand{\E}{\mathbb{E}}

\newcommand{\KL}{D_{\mathrm{KL}}}

\usepackage[utf8]{inputenc} % allow utf-8 input
\usepackage[T1]{fontenc}    % use 8-bit T1 fonts
\usepackage{hyperref}       % hyperlinks
\usepackage{url}            % simple URL typesetting
\usepackage{booktabs}       % professional-quality tables
\usepackage{amsfonts}       % blackboard math symbols
\usepackage{nicefrac}       % compact symbols for 1/2, etc.
\usepackage{microtype}      % microtypography
\usepackage{xcolor}         % colors

\usepackage{microtype}
\usepackage{graphicx}
\usepackage{subfigure}
\usepackage{booktabs} % for professional tables
\usepackage{bm}
\usepackage{subfigure}
\usepackage{multicol}
\usepackage{multirow}
\usepackage{subcaption}
\usepackage{soul}
\usepackage{float}
\usepackage{algorithm}
\usepackage{algorithmic}
\usepackage{etoolbox}
\AtBeginEnvironment{algorithmic}{\let\AND\relax}
\usepackage{enumitem}      % for \begin{enumerate}[noitemsep]
\usepackage{wrapfig}
\usepackage{multirow}

\makeatletter
\newcommand{\removeParBefore}{\ifvmode\vspace*{-\baselineskip}\setlength{\parskip}{0ex}\fi}
\newcommand{\removeParAfter}{\@ifnextchar\par\@gobble\relax}
\newcommand{\eq}{\begingroup\removeParBefore\endlinechar=32 \eqinner}
\newcommand{\eqinner}[2][aligned]{\endlinechar=32%
\begin{gather}\begin{#1}#2\end{#1}\end{gather}\endgroup\removeParAfter}
\makeatother

\DeclareDocumentCommand{\p}{ D<>{p} D<>{} r() }{
\def\content{#3}\patchcmd{\content}{|}{\;#2\vert\;}{}{}
\ensuremath{#1 #2(\content #2)}}

\DeclareDocumentCommand{\P}{ D<>{P} D<>{\big} r() }{
\def\content{#3}\patchcmd{\content}{|}{\;#2\vert\;}{}{}
\ensuremath{\operatorname{#1}#2(\content #2)}}

\DeclareDocumentCommand{\E}{ D<>{E} E{_}{{}} D<>{\big} r[] }{
\def\content{#4}\patchcmd{\content}{|}{\;#3\vert\;}{}{}
\ensuremath{\operatorname{#1}_{#2}#3[\content #3]}}

\DeclareDocumentCommand{\D}{ D<>{D} D<>{\big} r[] }{
\def\content{#3}\patchcmd{\content}{||}{\;#2\|\;}{}{}
\ensuremath{\operatorname{#1}\!#2[\content #2]}}

\NewDocumentCommand{\Nor}{ r() }{\P<Normal>](#1)}
\NewDocumentCommand{\Cat}{ r() }{\P<Cat>](#1)}
\NewDocumentCommand{\Bin}{ r() }{\P<Bin>](#1)}
\NewDocumentCommand{\Bet}{ r() }{\P<Beta>](#1)}
\NewDocumentCommand{\Ber}{ r() }{\P<Bernoulli>(#1)}
\NewDocumentCommand{\Dir}{ r() }{\P<Dir>(#1)}

\DeclareDocumentCommand{\KL}{ D<>{\big} r[] }{\D<KL><#1>[#2]}
\DeclareDocumentCommand{\H}{ D<>{\big} r[] }{\E<H><#1>[#2]}
\DeclareDocumentCommand{\I}{ D<>{\big} r[] }{\E<I><#1>[#2]}

\DeclareDocumentCommand{\pp}{ D<>{} r() }{
\ensuremath{\p<p_\theta><#1>(#2)}}
\DeclareDocumentCommand{\qp}{ D<>{} r() }{
\ensuremath{\p<q_\theta><#1>(#2)}}

\newcommand{\padspace}{\hspace{2em}}
\newcommand{\padspacenr}{\hspace{1em}}

\usepackage{mathtools}
\usepackage{amsthm}
\usepackage[capitalize,noabbrev]{cleveref}

\theoremstyle{plain}
\newtheorem{theorem}{Theorem}[section]
\newtheorem{proposition}[theorem]{Proposition}

\theoremstyle{definition}
\newtheorem{definition}[theorem]{Definition}

\theoremstyle{remark}

\usepackage[textsize=tiny]{todonotes}

\title{Efficient Generation of Diverse Cooperative Agents \\ with World Models}

\author{\name Yi Loo \email yi\_loo@mymail.sutd.edu.sg \\
      \addr Singapore University of Technology and Design (SUTD)
      \AND
      \name Akshunn Trivedi\thanks{Work done while author is at SUTD} \email akshunn@umich.edu \\
      \addr University of Michigan
      \AND
      \name Malika Meghjani \email malika\_meghjani@sutd.edu.sg\\
      \addr Singapore University of Technology and Design (SUTD) \\
}

\def\month{MM}  % Insert correct month for camera-ready version
\def\year{YYYY} % Insert correct year for camera-ready version
\def\openreview{\url{https://openreview.net/forum?id=XXXX}} % Insert correct link to OpenReview for camera-ready version

\begin{document}

\maketitle

\begin{abstract}
A major bottleneck in the training process for Zero-Shot Coordination (ZSC) agents is the generation of diverse partner agents via Population-based Training (PBT). Prominent PBT methods, such as  Cross-play Minimization (XPM), can be very computationally expensive and sample inefficient, as the training objective requires sampling multiple types of trajectories, in particular the need of Mixed-Play (MP) trajectories to mitigate self-sabotage in cross-play. Each partner agent in the population is also trained from scratch, despite all of the partners in the population learning policies of the same coordination task. In this work, we propose that simulated trajectories from the dynamics model of an environment can drastically speed up the training process for XPM methods. We introduce XPM-WM, a framework for generating simulated trajectories for XPM via a learned World Model (WM). We show XPM with simulated trajectories removes the need to sample Mixed-Play trajectories while still mitigating any self-sabotage. The learned WM also bootstraps the training of subsequent agents in the population. We show that these two factors combined gives XPM-WM significantly better sample efficiency as compared to other XPM methods and scales linearly to the size of the population. We show our proposed method can effectively generate partners with diverse conventions that match the performance of previous methods. Finally, we show that our proposed training paradigm of using learned world models to generate diverse population can be extended to other PBT methods beyond XPM.

\end{abstract}

\section{Introduction}
\label{sec:introduction}

% Intelligent cooperative agents hold a lot of promise in the field of artificial intelligence. A great number of multi-agent tasks in the real-world require intelligent decision-making involving some element of cooperation and coordination. Some prominent examples include intelligent drone swarms \citep{kouzeghar2023multi, huang2024collision},  assistive robots in healthcare \citep{feil2005defining,feil2011socially} and team-based games \citep{berner2019dota, bard2020hanabi,samvelyan2019starcraft}.
% An effective cooperative agent should generalize well to new partners and teammates in zero shot. However, this remains a challenging task for Reinforcement Learning (RL) agents, mainly due to the distributional shift of partner policies from training to those encountered in inference \citep{carroll2019utility}. Recently, t

The Zero Shot Coordination (ZSC) problem \citep{hu2020other} was proposed to evaluate a cooperative agent's ability to generalize to novel partners not encountered during training. Currently one of the most prominent way to train ZSC agents is through population-based training (PBT) \citep{lupu2021trajectory,strouse2021collaborating,zhao2023maximum,loo2023hierarchical,yulearning, lou2023pecan, rahman2024minimum}, where an ego agent is trained together with a population of diverse agents. Additionally, diverse agents also serve as holdout sets of novel partners to evaluate the performance of ZSC agents against novel partners.

 \begin{figure*}[t!]
\centering
\vspace{-5mm}

\subfigure{\includegraphics[width=\textwidth]{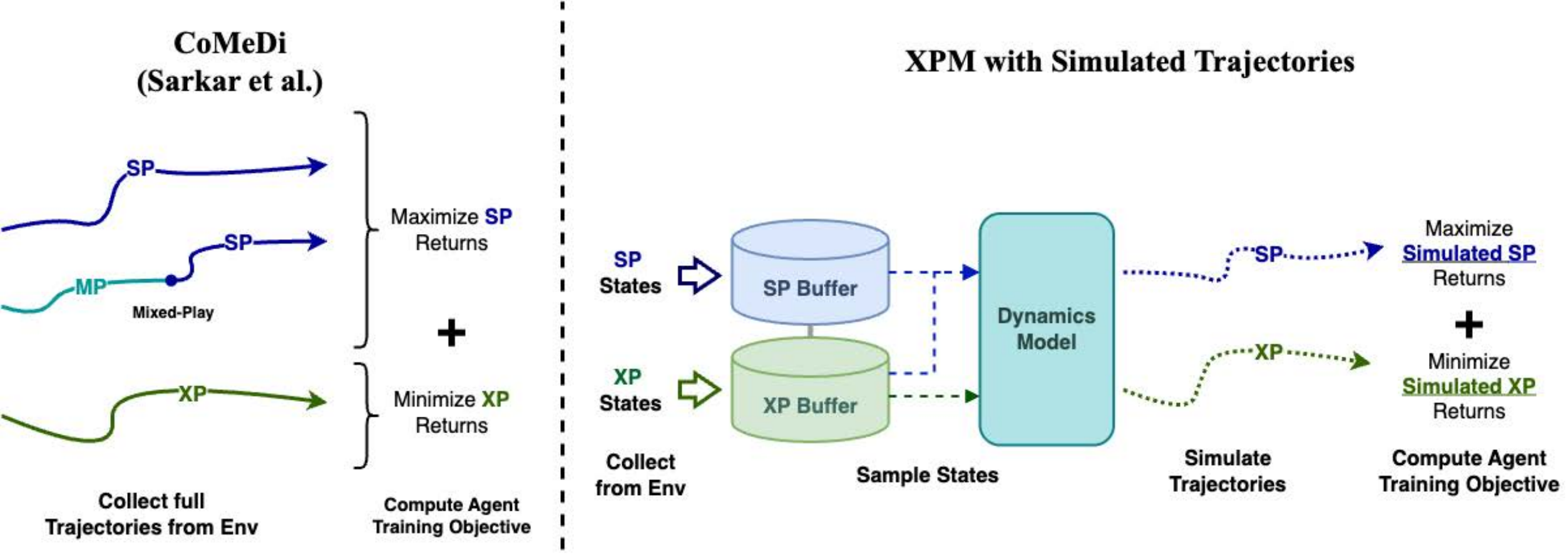}}

% \vspace{-5mm}

\caption{Diagram of the training process of XPM-Sim (right) compared to CoMeDi. Our key insight is that by expanding the starting states of SP simulated trajectories, we remove the need for Mixed Play trajectories while still producing policies that do not self-sabotage, resulting in a more sample efficient method.}
\label{fig:comedi_xpmsim_comp}

\vspace{-5mm}

\end{figure*}

Thus, generating a diverse set of policies is a crucial step in the training and evaluation of ZSC agents. However, PBT methods can be very computationally expensive and sample inefficient, as each agent within the population is trained from scratch \citep{lupu2021trajectory, zhao2023maximum, charakorn2023generating,sarkar2024diverse}, despite being trained on identical tasks. Furthermore, a prominent class of  diverse population generation methods uses cross-play minimization (XPM) \citep{charakorn2023generating,sarkar2024diverse} which require sampling additional types of trajectories outside of self-play trajectories. LIPO \citep{charakorn2023generating}, for example, requires sampling cross-play trajectories from every partner agent at every training step.
% making it both sample inefficient and computationally expensive. In addition to these two challenges, it is known that XPM can cause self-sabotage behavior. Definition of self-sabotage. Comedi solves the self-sabotage problem however, the sample in-efficiency and computation cost is still significantly expensive which is not suitable for scaling to multiple agents. s
Vanilla XPM is also known to cause agent to self sabotage in cross play \citep{cui2023adversarial}, CoMeDi \citep{sarkar2024diverse} proposes sampling mixed-play trajectories on top of self- and cross-play trajectories at every training iteration to mitigate self-sabotaging behavior, however this adds even more sample inefficiency to the training process. As a result, the amount of data (i.e. environment interactions) required for training each agent scales super-linearly with the number of existing partner agents in the population. Additionally, most previous PBT approaches also use on-policy methods \citep{yu2022surprising} to train each agent within the population, which is known to be less sample efficient than off-policy and model-based methods.

%Our approach .... which helps to overcome the self-sabotage problem while maintaining sample efficiency to keep the computation cost significantly lower. These features are highly desirable for scaling up.
As the task and environment complexity increases, it may to paramount to generate populations of larger sizes, hence the ability to efficiently scale up training of diverse agents is highly desirable. In this work, we present a more sample efficient method for generating diverse cooperative agents. We focus our method on addressing two key areas of sample inefficiency in current XPM methods: (i) the requirement to sample multiple types of trajectories and (ii) the lack of knowledge transfer between agents within the population due to each agent being trained from scratch. At the same time, we also would also like to ensure that our method retains the ability to generate agents that are meaningfully diverse and do not self-sabotage \citep{sarkar2024diverse, cui2023adversarial}. We propose that Model-Based Reinforcement Learning (MBRL) \citep{m2023model} could be used to improve sample efficiency in XPM methods. In particular, we show that, given an accurate model of the environment, we can derive a more sample efficient XPM objective by learning exclusively through simulated self- and cross-play trajectories. In addition, we empirically show that our proposed objective, XPM-Sim, removes the need for mixed-play trajectories while still producing agent policies that do not self-sabotage.

For a more general case where the model of the environment is unknown, we present XPM-WM, where a world model \citep{ha2018world,hafnerdream} is learned jointly with XPM agents. Our method first trains a world model to act as a strong prior to bootstrap the training of individual partner agents. We then train diverse agents with simulated self and cross-play trajectories generated by the world model with XPM-Sim. We validate XPM-WM on both fully observable and partially observable versions of the Overcooked AI environment \citep{carroll2019utility,gesslerovercookedv2}. We show that  XPM-WM scales linearly with the number of agents in terms of environment interactions. We then show that XPM-WM generates a population of comparable degree of diversity as other XPM methods. We also show that our proposed framework of using world models to bootstrap the training of diverse partner agents can be extended to other prominent non-XPM methods \citep{strouse2021collaborating, zhao2023maximum} with minimal modifications. Finally, we show that the generated agent population from XPM-WM can act as an effective population for ZSC ego agent training by evaluating the resultant ego-agent on novel agent populations as well as human partners.
% \vspace{-0.01\textheight}
To summarize, our contributions in this paper are as follows:
\vspace{-0.01\textheight}
\begin{enumerate}[noitemsep]
    \item We introduce XPM-Sim, a sample efficient objective for generating diverse agents via XPM by training on simulated trajectories. We empirically show that XPM-Sim removes the need for sampling mixed-play trajectories while preserving non self-sabotaging properties.
    \item For cases where the environment model is unknown, we propose XPM-WM, which extends XPM-Sim by utilizing world models to simulate self-play and cross-play trajectories.
    \item  We show that XPM-WM is more sample efficient than other XPM methods and scales linearly in terms of environmental steps with the number of agents in the population. 
    \item We show that agents generated via XPM-WM are effective as training partners for ZSC agents by validating the trained ego agents on novel agent partners as well and human partners.
    \item We show that our proposed framework of using world models to generate diverse partners can be extended to non-XPM PBT methods.
\end{enumerate}

\section{Related Work}
\label{sec:related_work}

\textbf{World Models:} World Models have been a popular branch of Model-Based Reinforcement Learning methods, most prominent of which are the Dreamer models \citep{hafnerdream,hafnermastering,hafner2023mastering}, which learns the state transitions in form of a Recurrent State Space Model (RSSM) and allow agents to learn via simulated trajectories. 
% World Models have been applied to many areas of intelligent control such as simulating entire game environments using large scale training data \citep{bruce2024genie,parker2024genie}, and in real-world scenarios such as robot control and navigation \citep{wu2023daydreamer,bar2024navigation} and autonomous driving \citep{wang2024drivedreamer, hu2023gaia}.
In the multi-agent context,  \cite{venugopal2024mabl} proposed MABL, which adapts the Dreamer model for a Multi-Agent Reinforcement Learning setting, in which agents in a team are trained jointly. MABL divides the world model into a hierarchical model consisting global and agent specific level. Our method does not explicitly separate the global and agent level representations at the architecture level but instead split the latent representations itself into global and agent-specific representations. \cite{toledo2024codreamer} and \cite{zeng2025efficient} explore utilizing world models to facilitate communication between agents in a MARL setting, specifically in the Centralized Training and Decentralized Execution (CTDE) setting. Our proposed approach focuses instead on the ZSC setting where agent are evaluated with novel partners cannot communicate with each other.

% Another related work using WM is by Prasanna et al. \citep{prasanna-rlc24a}, which proposed using world models for Zero-Shot Generalization of single-agent RL agents. They propose doing so by adding contextual information into world model to allow agents to generalize to out-of distribution tasks. Though we are also interested in the generalization to unseen partners in zero-shot, we instead employ world models as a means to efficiently generate new partner agents to train a robust ego agent that can do Zero shot Coordination.

% \input{figures_and_tables/fig_xpmwm_overview}

\textbf{Other Methods for Diverse Population Generation:}
TrajeDi \citep{lupu2021trajectory} uses Jensen-Shannon Divergence objective on policy trajectories and states of each partner agent to train a statistically diverse set of partner agents.   HSP \citep{yulearning} trains individual agent with hidden biased reward on top of the existing environmental rewards. The hidden rewards are randomly sampled for each agent based on sub-events within the environment. A best-response policy is then trained together with each biased agent and the biased-best response agent pair is then added to form the pool of partner agents. Similar to our method, ADVERSITY \citep{cui2023adversarial} also utilize a model-based method for cross-play minimization, specifically to learn an incompatible policy from a base policy in a partially observable environment of Hanabi. Rather than learning a world model, ADVERSITY instead learns a belief model \citep{hu2021off}, which models behavior of the partner agents as well as the true trajectory/state. ADVERSITY then uses the belief model to generate fictitious transitions similar to how World Model generates simulated trajectories. ADVERSITY also requires training multiple belief models for agent which is expensive to compute as the number of agents increases. Our method instead only learns a single World Model which is continuously fine-tuned over the training process of every agent.

\textbf{Alternate Approaches to ZSC/Ad-hoc Teamwork:}
GAMMA \citep{liang2024learning} uses a diverse population to train a VAE \citep{kingma2013auto} that generates diverse partners for ego agent training by interpolating between partner agent representations in latent space. GOAT \citep{chaudhary2025improving} builds upon GAMMA by adding an adversary that searches for adversarial agents in latent space of the VAE. Our proposed approach is complementary to both approaches as the world model framework can be used to efficiently generate the diverse population for VAE training. ROTATE \citep{wangrotate} proposes an open ended approach to AHT training that involves continuously generating partner agents that maximizes the per trajectory regret of the ego agent. Similar to XPM methods, ROTATE also requires sampling of SXP trajectories (trajectories that start with SP and switches to XP). Hence, our proposed framework of using world models to simulate SP trajectories from XP states is therefore compatible with ROTATE and can be used in conjunction with the approach. 
% \citet{jha2025cross} introduced CEC, an alternative to PBT that trains on diverse procedurally generated environments rather than partners. They demonstrate that training ego agents on procedurally generated environments allow for generalization to both novel partners and environments.
\section{Preliminaries}
\label{preliminaries}

\subsection{Zero-Shot Coordination (ZSC)}

The Zero-shot Coordination (ZSC) problem is typically formulated as a Decentralized  Partially Observable Markov Decision Process (Dec-POMDP), represented by the tuple
$(\mathcal{S}, O, \{\mathcal{O}^{i},\mathcal{A}^{i}\}_{i\in\mathcal{N}}, T, R,\gamma)$. $\mathcal{S}$ represents the state space of the environment. $\mathcal{N}$ represents the set of players in the environment, we focus on two-player tasks in this work hence in this case $\mathcal{N} = \{1,2\}$. $\mathcal{O}^{i}$ and $\mathcal{A}^{i}$ are the observation and action spaces of the $i$-th player respectively. The observation function $O: \mathcal{S} \rightarrow \mathcal{O}^1 \times \mathcal{O}^2$ generates a player specific observation of the state. We can write the joint action space as $\bm{\mathcal{A}}: \mathcal{A}^{1} \times \mathcal{A}^{2}$ and a joint action as $\bm{a} = (a^1,a^2)$. The transition function $T: \mathcal{S} \times \bm{\mathcal{A}} \times \mathcal{S} \rightarrow [0,1]$ represents the probability of reaching a subsequent state given a previous state and joint action. The reward function $R: \mathcal{S} \times \bm{\mathcal{A}} \rightarrow \mathbb{R}$ gives a real valued reward after each state transition and finally $\gamma$ represents the reward discount factor. 

A joint policy $\bm{\pi}: \pi^{1} \times \pi^{2}$ acts on the Dec-POMDP by producing (joint) action distributions, at every state $s \in \mathcal{S}$. Over a finite horizon $H$, the policy produces a joint trajectory, $\bm{\tau_{\pi}} = (s_0,\bm{a}_0,..., s_{H-1},\bm{a}_{H-1},s_{H})$, which can be further divided into player specific trajectories $\tau^{i}_{\pi} = (o^{i}_0,a^{i}_0,..., o^{i}_{H-1},a^{i}_{H-1},o^{i}_{H})$. The discounted return for a trajectory is defined as $G(\bm{\tau}) = \sum^{H-1}_{t=0}\gamma^{t} r(s_t,\bm{a}_t)$ and the expected return for a joint policy is defined as $J(\pi^1,\pi^2) = \mathbb{E}_{\bm{\tau} \sim (\pi^1,\pi^2)}[G(\bm{\tau})]$. The self-play (SP) return, $J_{SP}(\pi_A)$ of policy $\pi_A$ is when both players in the joint policy are identical copies, i.e. $\pi^1 = \pi^2 = \pi_A$. The cross-play (XP) return for two distinct policies, $\pi_A$ and $\pi_B$, is then defined as: $J_{XP}(\pi_A,\pi_B) \equiv J(\pi_A,\pi_B) + J(\pi_B,\pi_A)$.

The objective of ZSC is then to produce an ego-agent, represented by 
 policy $\pi^e$ that maximizes cross-play returns with policies drawn from unseen set of policies, $\Pi_{test}$. The ZSC objective is thus defined as $J_{ZSC}(\pi_e) = \max_{\pi^{e}}\mathbb{E}_{\scriptstyle{\pi^{\ast}\sim\Pi_{test}}}[J_{XP}(\pi^e,\pi^{\ast})]$. The original formulations of ZSC require the unseen policies, $\Pi_{test}$ to be also trained using the same algorithm as $\pi^e$ \citep{hu2020other}, but more recent work have expanded the formulation to include policies trained with other methods \citep{lucas2022any} or even human players \citep{li2024tackling, wang2024zsc}, which is more in-line with the formulation for Ad-hoc Teamwork as proposed by Stone et al. \citep{stone2010ad}. In this work we adopt the latter, broader definition of ZSC.

% \begin{equation}
%    J_{ZSC}(\pi_e) = \max_{\pi^{e}}\mathbb{E}_{\scriptstyle{\pi^{\ast}\sim\Pi_{test}}}[J_{XP}(\pi^e,\pi^{\ast})]
% \end{equation}

% Initial formulations of ZSC require the unseen policies, $\Pi_{test}$ to be also trained using the same algorithm as $\pi^e$ \citep{hu2020other}, but later works have expanded the formulation to include policies trained with other methods \citep{lucas2022any} or even human players \citep{li2024tackling, wang2023quantifying}, which is more in-line with the formulation for Ad-hoc Teamplay as proposed by Stone et al. \citep{stone2010ad}. In this work we adopt the latter, broader definition of ZSC.

% \input{figures_and_tables/tab_xpm_sim_results}

\begin{figure*}[t!]
\vspace{-4mm}

% \begin{wrapfigure}{L}{0.5\textwidth}
\centering
\includegraphics[width=\textwidth]{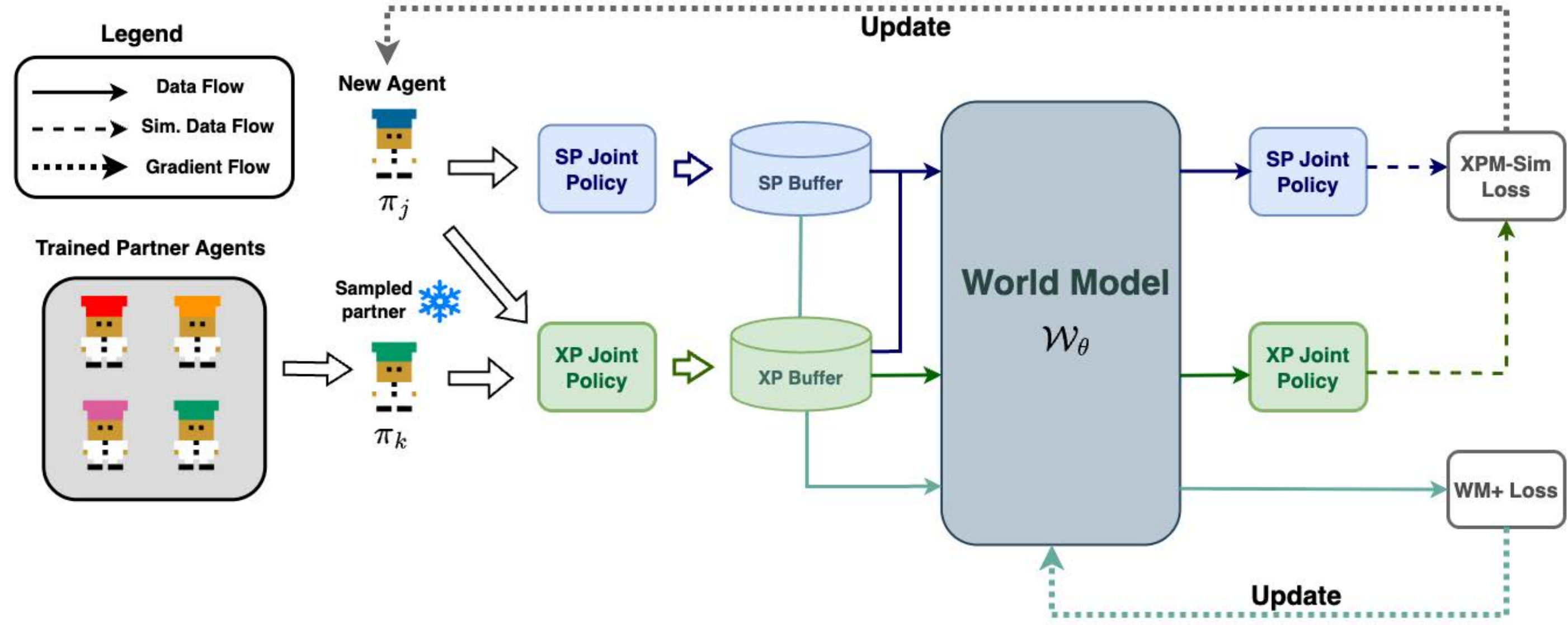}

% \vspace{-3mm}

\caption{Diagram of the training process for XPM-WM. For the first phase, only the self-play (in blue) portion of the diagram is used and the agent is trained with just the objective SP. During the training of subsequent agents using XPM-Sim, the parameters of the partner agent is kept frozen while we continuously update the World Model.}
\label{fig:popdoverview}

\vspace{-4mm}

\end{figure*}
% \end{wrapfigure}

\subsection{Population Based Training with Cross-Play Minimization }

% Population-based training is where the ego agent, $\pi^e$ is trained together with a diverse set of partner policies $\Pi_{train}$.
% The idea is that a sufficiently diverse $\Pi_{train}$ that covers a variety of conventions \citep{shih2020critical} allows the ego agent to learn a robust policy that encapsulates the best-response (BR) policies to each member of $\Pi_{train}$ and thus generalize to unseen policies in $\Pi_{test}$. A set of diverse partner policies, trained separately from $\Pi_{train}$ can also act as surrogate for $\Pi_{test}$ to evaluate the cooperative capabilities of the ego agent.

% \subsection{Generating A Diverse Agent Population via Cross-Play Minimization}

% \textbf{Cross-Play Minimization:}
We now formalize the problem of generating a diverse population of policies for a 2-player cooperative task. We are interested in generating a set of $M$ agents each represented by policy $\bm{\pi}_j$ , $\mathcal{D} = \{\bm{\pi}_{1},\bm{\pi}_{2},...,\bm{\pi}_{m}\}$ that are diverse in their behavioral conventions. A convention is defined as a joint policy $\bm{\pi_{j}} = \pi^{1}_{j} \times \pi^{2}_{j}$. Two policies, $\bm{\pi_{A}}$ and $\bm{\pi_{B}}$ are said to have compatible conventions if their joint cross-play objective is at least as high as the self-play objective \citep{charakorn2023generating}. Thus, a straightforward way to generate diverse policies is to ensure the pairwise cross-play objective for each distinct pair of policies in the population is divergent from the individual policies' self-play objectives. This leads to a broad class of XPM methods, where the agents are trained to maximize $J_{SP}$ while simultaneously minimizing the pairwise $J_{XP}$ objective with respect to other agents in $\mathcal{D}$. 

% \eq{
%     J_{XP}(\bm{\pi_{A}, \bm{\pi_{B}}}) \geq \frac{(1 - \epsilon)}{2}(J_{SP}(\bm{\pi_{A}}) + J_{SP}(\bm{\pi_{B}}))
% }

% Thus, a straightforward way to generate diverse policies is to ensure the pairwise cross-play objective for each distinct pairs of policies in the population is significantly lower than the individual policies' self-play objectives,

% \eq{
%     J_{XP}(\bm{\pi}_{j}, \bm{\pi}_{k}) < \frac{(1 - \epsilon)}{2}(J_{SP}(\bm{\pi}_{i}) + J_{SP}(\bm{\pi}_{k})),  
%     \forall \bm{\pi}_{j}, \bm{\pi}_{k} \in \mathcal{D}, \bm{\pi}_{j} \neq \bm{\pi}_{k}
% }

% This leads to a broad class of Cross-Play Minimization (XPM) methods, where the agents are trained to maximize $J_{SP}$ while simultaneously minimizing the pairwise $J_{XP}$ objective with respect to other agents in $\mathcal{D}$. 

\textbf{LIPO:}
 LIPO  proposed by Charakorn et al. \citep{charakorn2023generating} presents a straightforward method for XPM training. Specifically for given policy $\bm{\pi}_j \in \mathcal{D}$, the LIPO method optimizes the following objective:

\eq{
    J_{LIPO}(\bm{\pi}_{j}) =  J_{SP}(\bm{\pi}_{j}) - \lambda_{\scriptscriptstyle{XP}} \max_{\bm{\pi}_{k} \in \mathcal{D}\setminus\bm{\pi}_{j}}J_{XP}(\bm{\pi}_{j},\bm{\pi}_{k})
\label{lipo}
}

Although LIPO is effective in generating populations that have low cross-play returns, the choice to train all agents concurrently could result in training instabilities. Furthermore, recent work \citep{cui2023adversarial,sarkar2024diverse} demonstrated that the base XPM training objective could result in policies that learn to perform handshakes and self-sabotage in cross-play. 
% Individual policies within $\mathcal{D}$ could potentially learn certain arbitrary sequence of actions in self-play that allows agents to identify whether the partner agent is a SP policy or another agent in $\mathcal{D}$. If the agent identitifies its partner is indeed a separate agent, it can then learn to self-sabotage and give the false premise that it learns incompatible policies by displaying low cross-play returns with said other agent.

% \input{figures_and_tables/fig_mpecomp}

\textbf{CoMeDi:}
CoMeDi \citep{sarkar2024diverse}, was proposed as an augmentation on top of the LIPO objective to mitigate the issue of handshakes/self-sabotaging. Specifically, CoMeDi proposes adding a Mixed-Play (MP) term to the overall training objective. A MP trajectory is generated by randomly sampling actions from either SP or XP until a random timestep, $t^{\ast}$ before horizon $H$, before switching to SP for the remainder of the episode. The trajectory before $t^{\ast}$ is discarded and the remaining trajectory segment is trained with the self-play objective. The full CoMeDi objective then becomes

\eq{
\begin{split}
    J_{CoMeDi}(\bm{\pi}_{j}) =  (J_{SP}(\bm{\pi}_{j}) + \lambda_{\scriptscriptstyle{MP}}J_{MP}(\bm{\pi}_{j})) 
    - \lambda_{\scriptscriptstyle{XP}} \max_{\bm{\pi}_{k} \in \mathcal{D}\setminus\bm{\pi}_{j}}J_{XP}(\bm{\pi}_{j},\bm{\pi}_{k})
\end{split}.
}

Additionally, CoMeDi also proposed to train each agent sequentially rather than concurrently, where agents are trained one at a time while keeping previous agent's policies frozen, resulting in a more stable training process.

% \subsection{Limitations of Cross-Play Minimization}

% We identify that current XPM methods are sample inefficient in two key areas: (a) the need to sample additional XP trajectories from each previous partner at every training iteration. This is to compute the XP objective but also to determine the maximum XP partner. (b) The need to sample MP trajectories to mitigate self-sabotage and (c) the lack of knowledge transfer between agents in $\mathcal{D}$ during training. In the following sections, we outline our proposed framework of using Model-Based Reinforcement Learning (MBRL) approach to address these limitations. We first demonstrate how simulated trajectories from a dynamics model can increase the sample efficiency of XPM. We then outline our proposed approach, XPM-WM as a more generalizable method where simulated trajectories are generated via a learned World Model (WM). We also show that utilizing WMs in XPM also amortizes the cost of training subsequent agents by providing a strong prior to bootstrap the training process of subsequent agents.

\subsection{Other Population Based Training Methods}

\textbf{FCP:}
Fictitious Co-Play (FCP) \citep{strouse2021collaborating} augments existing SP populations by adding past training checkpoints of agents within the partner population to simulate lower-skilled partners.

\textbf{MEP:}
Maximum Entropy Population-based Training (MEP) \citep{zhao2023maximum} enforces population diversity in a partner population by maximizing the Population Entropy within the population, $\mathcal{H}(\bar{\pi}(\cdot|s_t))$. In practice, each agent is trained with an augmented training reward $r^{MEP}(s_t, \mathbf{a}_t)  =  r(s_t, \mathbf{a}_t) + \alpha log(\bar{\pi}(\mathbf{a}_t|s_t)) $, where $\bar{\pi}(a_t|s_t)$ is the mean policy, $\bar{\pi}(a_t|s_t) = \frac{1}{m}\sum_{\pi_j \in \mathcal{D}}\pi_j(a_t|s_t)$.

\section{Method}
In the following sections, we outline how our proposed framework of using Model-Based Reinforcement Learning (MBRL) can address the issue of sample inefficiency and poor scaling in current PBT and in particularly, XPM methods. More specifically, we would to like address two aspects of current XPM methods that contributes to the sample inefficiency: (i) the need to sample additional XP trajectories from each previous partner at every training iteration. This is to compute the XP objective but also to determine the maximum XP partner. (ii) The need to sample MP trajectories to mitigate self-sabotage.

\subsection{Cross-Play Minimization with Simulated Trajectories}
\label{sec:xpm_sim}

% In the following sections, we outline how proposed framework of using Model-Based Reinforcement Learning (MBRL) can address the sample inefficiency of current PBT and in particularly, XPM methods. More specifically, we would like address two aspects of current XPM methods that contributes to the sample inefficiency: (i) the need to sample additional XP trajectories from each previous partner at every training iteration. This is to compute the XP objective but also to determine the maximum XP partner. (II) The need to sample MP trajectories to mitigate self-sabotage.

% \input{figures_and_tables/fig2_popdoverview}

For a given Dec-(PO)MDP, if the forward dynamics $T(s,\mathbf{a})$ and $R(s,\mathbf{a},s')$ are known, or can be accurately estimated via a model $F$, we can generate a simulated trajectory from any state $s \in \mathcal{S}$. Specifically, we define $\hat{\mathbf{\tau}}^{\bm{\pi}}_{F}(s) = (s, \mathbf{a}_0, \hat{s}_1, \mathbf{a}_1,...,\hat{s}_{H' -1}, \mathbf{a}_{H' -1}, \hat{s}_{H'})$ as a simulated trajectory with simulation horizon $H'$ that is generated using model $F$ starting from state $s$ with joint policy $\bm{\pi}$, where $\hat{s}_t$ is a simulated state at time $t$. Given two distinct policies, $\pi_j$ and $\pi_k$ in $\mathcal{D}$, we can generate SP and XP simulated trajectories, $ \hat{\tau}^{SP} = \hat{\mathbf{\tau}}^{(\pi_j,\pi_j)}_{F}(s)$ and $\hat{\tau}^{XP} = \hat{\mathbf{\tau}}^{(\pi_j,\pi_k)}_{F}(s)$. We are then able train a (joint) policy over simulated trajectories by maximizing the expected simulated returns, $\widehat{J}(\bm{\pi}) = \mathbb{E}_{\bm{\hat{\tau}}_F \sim (\bm{\pi}), s \in \mathcal{S}}[G(\bm{\hat{\tau}}_F(s))]$. Note that the current objective allows trajectories to be simulated from any starting state in $\mathcal{S}$. However, for XPM training we restrict the set of starting states of simulated SP and XP to the set of \textit{reachable} states which we define below:
% \eq{
% \hat{\mathbf{\tau}}^{\bm{\pi}}_{F}(s) = (s, \mathbf{a}_0, \hat{s}_1, \mathbf{a}_1,...,\hat{s}_{H' -1}, \mathbf{a}_{H' -1}, \hat{s}_{H'})
% }

% \eq{
% \widehat{J}(\bm{\pi}) = \mathbb{E}_{\bm{\hat{\tau}}_F \sim (\bm{\pi}), s \in \mathcal{S}}[G(\bm{\hat{\tau}}_F(s))]$$
% }

% Note that the current objective allows trajectories to be simulated from any starting state in $\mathcal{S}$. However, for XPM training we restrict the set of starting states of simulated SP and XP to the set of \textit{reachable} states which we define in the following:

\begin{definition}
For a two-player Dec-MDP given a joint policy $\bm{\pi} = (\pi^1,\pi^2)$ the set of reachable states, $\mathcal{S}^{\bm{\pi}} \subseteq \mathcal{S}$ consist of any \textit{reachable} state under joint policy $\bm{\pi}$ from starting state $s_0 \sim p(s_0)$ within horizon $H$.
\end{definition}

For each policy in $\mathcal{D}$, we can also define the reachable self-play states, $\mathcal{S}^{j}_{SP} = \mathcal{S}^{\bm{\pi_j}}$ and reachable cross-play states, $\mathcal{S}^{j,k}_{XP} = \mathcal{S}^{(\pi_j,\pi_k)}$. This then allows us to define $\widehat{J_{SP}}(\pi^j)$ and $\widehat{J_{XP}}(\pi^j, \pi^k)$ and arrive at a similar XPM objective as in Equation \ref{lipo}. However, without the addition of Mixed-Play objective, policies may still establish handshakes and learn to self-sabotage. We propose that we are able to remove the requirement of the Mixed play objective by simply expanding the starting states of simulated SP trajectories to include reachable XP states.

\begin{proposition}
\label{sim_prop}
Training a cooperative agent policy on simulated SP trajectories, $\mathbf{\hat{\tau}}^{\pi_{SP}}_F(s_{xp})$ generated via dynamics model $F$ from any cross-play state, $s_{xp} \in \mathcal{S}_{XP}$ is able to mitigate handshake and self-sabotage behavior in an XPM agent population, similar to training on Mixed-Play trajectories.
\label{xpm_sim_prop}
\end{proposition}

 We experimentally show this proposition to be true in the following subsection. We can now construct a XPM training objective with simulated trajectories, XPM-Sim:

% {\small
\eq{
\begin{split}
    J_{XPMSim}(\bm{\pi}_{j}) = &\max_{\pi_{j}} \widehat{J_{SP}}(\bm{\pi}_{j})  - \lambda_{\scriptscriptstyle{XP}} \max_{\pi_{k} \in \mathcal{D}\setminus\pi_{j}}\widehat{J_{XP}}(\pi_{j},\pi_{k}), \\
    \text{where}~\widehat{J_{SP}}(\bm{\pi}_{j}) = &~\mathbb{E}_{\bm{\hat{\tau}}_F \sim (\bm{\pi_j}), s \in \{\mathcal{S}^j_{SP},\mathcal{S}^{j,l}_{XP}\}}[G(\bm{\hat{\tau}}_F(s))], 
    ~~\pi_l \in \mathcal{D} \setminus \pi_j, \\
    \text{and}~\widehat{J_{XP}}(\pi_{j}, \pi_{k}) = &~\mathbb{E}_{\bm{\hat{\tau}}_F \sim (\pi_{j}, \pi_{k}), s \in \mathcal{S}^{j,k}_{XP}}[G(\bm{\hat{\tau}}_F(s))].
\end{split}
\label{xpm-sim}
}
% \vspace{-mm}

Note the above formulation allows for simulated SP trajectories to start from reachable SP states and reachable XP states with \textit{any} agent in $\mathcal{D}$ and not just the maximum XP partner.
XPM-Sim provides a significantly more sample efficient training process over CoMeDi as (i) XPM-Sim does not require sampling Mixed-Play trajectories from the environment at every training iteration, (ii) even though the sampling SP and XP trajectories are still required, they can be reused by storing them in SP and XP replay buffers and (iii) the maximum XP partner can be determined by looking up the returns of latest trajectories in the replay buffer rather than resampling at every iteration. We show an outline of XPM-Sim in Figure \ref{fig:comedi_xpmsim_comp} and contrast it with CoMeDi.

% \input{figures_and_tables/fig2_popdoverview}

% \subsection{Toy Experiment in Multi-Particle Environment}
\subsubsection{Empirical Analysis of Self-Sabotage Behavior}
\label{toy_exp}

We will now empirically show Proposition \ref{sim_prop} to be true via a simple toy experiment. We specifically want to compare the amount of self-sabotaging behavior in agents trained using XPM-Sim with other XPM methods, LIPO and CoMeDi. As it is challenging to define what constitutes self-sabotaging behavior in more complex cooperative environments, we conduct this experiment in the simple Multi Particle Environment (MPE) \citep{lowe2017multi}. Specifically, we adopt the Multi Particle Point Mass Rendezvous (MPPMR) task proposed by \citep{charakorn2023generating}, where agents, represented as point particles are tasked to rendezvous at any of the four evenly spaced checkpoints in the environment.

To adapt this task for checking self-sabotage behavior, we add a fixed boundary to the environment. Any agent particle that leaves the boundary of the environment will result in the episode to prematurely terminate. Hence, an agent that has learned to self-sabotage in cross-play can learn to immediately go out of environment boundary and thus minimizing the cross-play returns without learning to be meaningfully incompatible in terms of conventions with its partner agent.

As the forward dynamics of the MPE environment are determined by a series of linear dynamical equations, we can deterministically compute the forward dynamics of the MPPMR task from all starting states, which in turn allows us to simulate self and cross-play trajectories for XPM-Sim. We train a total of $M=8$ agents for each method and look at the number of instances where each agent in the population exhibits self-sabotaging behavior, which we determine by the fraction of early terminations in cross-play. 

 \textbf{Results:}
 We present the results in Table \ref{tab:mpe_ss}. Our results show XPM-Sim exhibits significantly less self-sabotage behavior than LIPO. More notably, XPM-Sim also exhibits less self-sabotage behavior as compared to CoMeDi. This suggests that XPM-Sim is more effective at mitigating self-sabotage behavior as XPM-Sim simulates new trajectories from every real cross and self-play states. This essentially allows the agent to simulate (pseudo) mixed-play from every state of cross-play trajectories, whereas in CoMeDi mixed-play trajectories only utilizes one start state sampled at timestep $t^*$ per trajectory while discarding every preceding state. Finally we also show that XPM-Sim achieves comparable level of self-play performance with respect to CoMeDi and LIPO.

\begin{table*}[t!]
    \centering
    \caption{Results of the MPPMR environment. The first 8 columns indicate the proportion of self-sabotage behaviors when in Cross-Play with preceding agents. E.g. Agent 4 will be paired with agents 1,2, and 3, with 1.0 indicating every episode containing self-sabotage behavior with 0 indicating None. SP indicates the average SP rewards across all 8 agents and XP indicates the average cross-play rewards.}
    \small{	
    \begin{tabular}{c|ccccccccc|cc}
        \toprule
        % \multirow{2}{*}{Layout}& \multicolumn{2}{c|}{$H_{proxy}$} & \multicolumn{2}{c}{ diverse SP} \\
          & $1$   & $2$ & $3$& $4$& $5$& $6$& $7$& $8$  & Average & SP $\uparrow$& XP $\downarrow$\\
        \midrule
        LIPO  &$0.0$  &  $1.0$  &  $1.0$  & $0.33$   & $0.5$   & $0.25$   &  $0.98$  &   $0.65$ & $0.59$  & $2.12$ & $-130.65$\\
        CoMeDi  & $0.0$   & $0.0$   & $0.0$   & $0.0$   & $0.0$   & $0.0$ & $0.175$   & $0.57$ & $0.09$ & $30.67$ & $-114.01$\\
        XPM-Sim (Ours) & $0.0$   & $0.0$   & $0.0$   & $0.0$   & $0.0$   & $0.0$ &  $0.09$   & $0.0$ & $\mathbf{0.011}$ & $21.50$ & $-114.50$\\
 
        \bottomrule
    \end{tabular}}

    \label{tab:mpe_ss}
    \vspace{-2mm}

\end{table*}

% We also include qualitative results from the three methods in Figure \ref{fig:mpe_qualitative}. We plot the trajectories of two agents in $\mathcal{D}$ in the MPPMR environment during cross-play from all three XPM methods. We show the agents trained under LIPO learns to go out-of-bounds while agents trained with CoMeDi and XPM-Sim stay within bounds and converge to separate landmarks. 

\subsection{Diverse Population Generation with World Models}
\label{sec:xpm_wm}

Although XPM-Sim is more efficient for generating diverse agents, the requirement for the full dynamics model, $F$ to be known could be impractical for more complex cooperative environments and tasks. In cases where the $F$ is not easily obtainable, we propose to instead estimate $F$ using a learned World Model \citep{ha2018world}.

We introduce XPM-WM, a method for generating diverse agents with XPM with simulated trajectories generated via a learned world model. XPM-WM consist of two training phases, the first phase involves learning a world model $\mathcal{W}_{\theta}$, parameterized by $\theta$ together with an initial SP agent $\bm{\pi}_{0}$. The trained world model is then used to bootstrap the training process of subsequent $M$ agents in $\mathcal{D}$ using the XPM-Sim training objective outlined in Equation \ref{xpm-sim}. On top of the increase in sample efficiency provided by XPM-Sim, using a World Model also makes the XPM training process more efficient by providing a strong prior in form of learned state representation from $\mathcal{W}_\theta$, effectively transferring knowledge about the environment across agents, rather than training each of them from scratch.

We will now describe in detail the training process for XPM-WM. We first describe how we adapt the Dreamer \citep{hafnermastering} world models to a two player cooperative task and self-play training. We then describe the training process for the world model. Finally, we describe how we train agents using XPM-Sim with the learned model. We show an overview of XPM-WM in Figure \ref{fig:popdoverview}.

% We now outline our proposed training framework for efficient generation of diverse agents using World Models. Our proposed method is as follows: we first train an initial world model $\mathcal{W}_{\theta}$, parameterized by $\theta$ together with the first agent $\bm{\pi}_{1}^{\phi_1}$ parameterized by $\phi_1$ in self-play. We then train subsequent agents $\bm{\pi}_2,...,\bm{\pi}_M$ in a sequential manner with cross-play minimization on top of the learned world model $\mathcal{W}_{\theta}$. In the following subsection we will outline the details of our proposed training method.

% \input{figures_and_tables/fig2_popdoverview}

\subsubsection{World Models for Self-Play Training}
\label{popd_des}
We adopt the Dreamer (v3) \citep{hafnermastering,hafner2023mastering} architecture as the base of the world model and make propose additional modifications for multi-agent self-play and cross-play training. The main components of the World Model include the encoder, recurrent state model, decoder, reward predictor, and continue predictor heads:

{\small{
\eq{
\begin{alignedat}{8}

& \text{Recurrent state model:}     &~~& h_t      &\ =  &\ f_\theta(h_{t-1},z_{t-1},a^{1}_{t-1}, a^{2}_{t-1}) ~~
&& \text{Transition predictor:}  \padspacenr & \hat{z}_t   &&\ \sim &\ \pp(\hat{z}_t | h_t) \\
& \text{Encoder:}  \padspacenr && z_t      &\ \sim &\ \qp(z_t | h_t,o^{1}_t,o^{2}_t) 
&& \text{Reward predictor:}    \padspacenr & \bm{\hat{r}}_t   &&\ \sim &\ \pp(\hat{r}_t | h_t,z_t) \\
& \text{Decoder:}    \padspacenr && \hat{o}^{i}_t   &\ \sim &\ \pp(\hat{o}^{i}_t | h_t,z^{i}_t), i \in \{1,2\} 
&& \text{Continue predictor:}   \padspacenr & \hat{\gamma}_t &&\ \sim &\ \pp(\hat{\gamma}_t | h_t,z_t).

\end{alignedat}
}}}

To facilitate a multi-agent two-player setup, we propose to modify the recurrent state model to take the previous joint actions of both players, $a^{1}_{t-1}$ and $a^{2}_{t-1}$. Similarly, the encoder model also takes the joint observations $o^1_t$ and $o^2_t$ and outputs a distribution of the latent representation of the current state. 

\textbf{Player-specific Latent Representation:}
We propose a decoder model that reconstructs player specific observations from the player-specific state representation $z^i_t$. To obtain $z^i_t = (z^{joint}_t, z^{p_i}_t)$, we first partition the state representation $z_t$ into parts, which consists a joint representation, $z^{joint}_t$ and player-centric representations, $z^{p_i}_t$. $z^{joint}_t$ represents global information in the state that are visible to all players (i.e. positions of certain objects) whereas $z^{p_i}_t$ represents player-specific information about the state (i.e. current position of the player). In practice we partition $z_t$ into $z^{joint}_t$ and by the number of discrete distributions. Specifically, out of the $N$ discrete distributions in $z_t$ , we set the first $K< N$ distributions as $z^{joint}_t$ and the remaining $(N - K)$ distributions are equally divided into $z^{p_i}_t$, and each $z^{p_i}_t$ would then consist of $\frac{(N - K)}{2}$ distributions. 

\textbf{Event-based Rewards:} We also propose a reward predictor that predict event-based rewards \citep{yulearning} over scalar rewards. Specifically, the reward predictor predicts a vector, where each column is a binary value indicating if an event or sub-task within the environment has been executed by a player $i$ in the current state $t$. In practice, we model the reward predictor as a $r$-dimensional Bernoulli distribution, where $r$ is the product of the number events and the number of players. For learning the policy in simulation, we compute the scalar reward as a linear function of the predicted event vector with a constant weight $w^r$. We find that this change greatly helps the World Model to effectively learn the transitional dynamics of the 2-player cooperative environment as it serves as a form of inductive bias that allow the model to tie player specific actions to the latent representation. We provide more details of the events specific to the environments we used in Supplementary Material.

% \textbf{Transition and Continue Predictor:}
% The transition predictor and continue predictor remains unchanged from the Dreamer model. Where the transition predictor predicts the next latent state without access to the full state, or in this case joint observation information. Finally, the continue predictor predicts the termination condition of the currently state.

\textbf{World Model Loss:}
For the initial phase of XPM-WM, we adopt Dreamer objective to train all the components in the World Model, which consists of the reconstruction, reward and continue predictor log loss, together with the KL Divergence term between $q_\theta(z_t)$ and $p_\theta(\hat{z}_t)$ \citep{hafnermastering}.

{\footnotesize{
\eq{
\begin{split}
  \mathcal{L}_{\theta} =~ & \mathbb{E}_{q_\theta}\bigr[\sum\nolimits^{H}_{t=1}[\sum\nolimits^{2}_{i=1}-ln p_\theta(o^i_t|h_t,z^i_t)] -ln p_\theta(\bm{r}_t|h_t,z_t) -ln p_\theta(\gamma_t|h_t,z_t) + \beta KL[q_\theta(z_t|h_t,\bm{o_t})||p_\theta(\hat{z}_t|h_t)]\bigr]
\end{split}
\label{wm_loss}
}}}

% Finally, we also employ KL balancing \citep{hafnermastering} and free nats \citep{hafner2023mastering} for the KL loss
% term as introduced in later versions of the Dreamer model.

\textbf{Actor and Critic:}
The actor takes in the player specific latent representation $\hat{z}^i_t$ and outputs the player action distribution. The joint actions are then fed back to the World Model to simulate future parts of the trajectory. For the Critic, we employ the Centralized Training and Decentralized (CTDE) \citep{kraemer2016multi,rashid2020monotonic} approach popular in MARL methods and train a centralized critic for each agent pair. The centralized critic take the full state representation $\hat{z}_t$ as input and outputs a distribution estimating the returns of simulated trajectories. Specifically for each agent $j$ in $\mathcal{D}$, we have:

% \eq{
% \begin{alignedat}{4}

% & \text{Actor:}    \padspace && a^i_t      \sim  \pi_{\phi^j}(a^i_t|\hat{z}^i_t), i \in \{1,2\} \\
% & \text{Centralized Critic:}  \padspace && v_{\psi^j_j}(\hat{z}_t)       \approx G(\bm{\hat{\tau}}_{\mathcal{W}_\theta}(s_t)).

% \end{alignedat}}

\eq{
\begin{alignedat}{9}

& \text{Actor:}        \padspace && a^i_t            \sim   \pi_{\phi^1}(a^i_t|\hat{z}^i_t), i \in \{1,2\} \padspace 
& \text{Centralized Critic:}   \padspace && v_{\psi^j_j}(\hat{z}_t)       \approx G(\bm{\hat{\tau}}_{\mathcal{W}_\theta}(s_t)).

\end{alignedat}}

Where $\phi^j$ are parameters of the actor, $\psi^j_j$ the parameters of the centralized SP critic and $\psi^j_k$ are parameters for the centralized XP critic for joint policy $(\pi_j,\pi_k)$. For an agent population of size $M$ we train a total of $\frac{M(M+1)}{2}$ critics as we train each agent sequentially.

\subsubsection{Training Subsequent Agents with a Learned World Model}

For the second phase of XPM-WM's training process, we train each subsequent agent in $\mathcal{D}$ sequentially using the XPM-Sim objective. We also continue fine-tuning the learned World Model, $\mathcal{W}_\theta$ with real cross and self-play trajectories. As in XPM-Sim, we maintain a separate SP and XP replay buffer and sample real trajectories to update the World Model. As the actor parameters of the previous agents are fixed, we would like to update the World Model such that its latent distribution does not drift away from that of the proceeding agents, especially for the set of XP reachable states $\mathcal{S_{XP}}$. This is so that proceeding agents will produce consistent actions to generate XP trajectories. Hence we add an additional KL Divergence to the world model training loss,

\eq{
\begin{split}
    \mathcal{L}^{WM+}_{\theta^j} = \mathcal{L}^{WM}_{\theta^j}+ \gamma KL[q_{\theta^j}(z_t|h_t,\bm{o_t})||q_{\theta^k}(z_t|h_t,\bm{o_t}))],  ~\bm{o_t} \sim (\pi_j,\pi_k)
\end{split}

\label{wm_loss_popd}
}

where $\mathcal{L}^{WM}$ is world model loss as in  Equation \ref{wm_loss}, $\theta^j$ are the parameters of the world model during the training process of agent $j$ and $\theta^k$ the (fixed) parameters of the world model during the training process of preceding agent $k$. After the training for agent $j$ is complete, we add WM $\mathcal{W}_{\theta^j}$ and policy $\pi_j$ to the pool of partner agents.

\subsubsection{Generating Non-XPM Populations with World Models}
\label{sec_mep_fcp}

Beyond XPM, we also extend the world model training framework for augment  prominent non-XPM Population Based Training methods FCP \citep{strouse2021collaborating} and MEP \citep{zhao2023maximum}. For FCP, we directly apply World Model training by first training a world model and train the subsequent agents on the learned WM using the SP training objective.

For MEP, we convert the existing MEP training framework where the population is trained jointly to a sequential population training framework. Specifically when training agent $j$-th, we modify the MEP reward to $r^{MEP-WM}(s_t, \mathbf{a}_t)  =  r(s_t, \mathbf{a}_t) - \alpha log(\bar{\pi}^\prime(\mathbf{a}_t|s_t)) $ where $\bar{\pi}(a_t|s_t)$ is the mean policy of the agents in the population that are trained so far, $\bar{\pi}^\prime(a_t|s_t) = \frac{1}{j}\sum^{j}_{k=0}\pi_k(a_t|s_t)$. We can then apply the World Model training framework to train the MEP agents sequently in with the SP objective. In the following sections we refer to the WM-augmented versions of FCP and MEP as FCP-WM and MEP-WM.
\section{Experiments}

\label{exp_sec}
In this section we provide experimental evaluation of XPM-WM. In Section \ref{efficiency_exp} we show the sample efficiency gains of XPM-WM compared to other XPM methods. In section \ref{sp_xp_score}, we demonstrate the diversity of the generated agents via the average SP and pair-wise XP scores of the generated agent populations. In Sections \ref{ego_exp} and \ref{sec:oc_v2}, we  further evaluate the diversity of the agents generated via XPM-WM on a full ZSC setting by training an ego agent with the population and evaluate its performance on a holdout set of partners. We then evaluate the ego-agents trained with FCP-WM and FPC-MEP in Section \ref{fcp_score}. In Section \ref{user_studies} we extend the ZSC setting to real human partners and finally in Section \ref{diversity_analysis}, we conduct a qualitative analysis on how well the diverse population covers different conventions present in human partners. We also include more detailed results in the Supplementary Material as well as additional ablation studies of the effects of the vectorized reward and the dynamics model on the training efficiency.

\textbf{Environment:}
We evaluate XPM-WM on the popular Overcooked AI cooperative environment \citep{carroll2019utility}. Two players control two separate cooks in the environment and are tasked to deliver as many soups as possible in a given time frame. To be successful, agents have to subdivide tasks among themselves, ensure that they do not impede their partner's progress and help them where necessary. We conduct our experiments on the five baseline layouts in Overcooked AI. We also evaluate XPM-WM on Overcooked v2 \citep{gesslerovercookedv2}, which provides a partially observable version of the base overcooked layouts.

\textbf{Baselines:}
We compare XPM-WM with two other cross-play minimization approaches, LIPO \citep{charakorn2023generating} and CoMeDi \citep{sarkar2024diverse}, as well as two non-XPM PBT methods, FCP \citep{strouse2021collaborating} and MEP \citep{zhao2023maximum}. All baselines are implemented using MAPPO \citep{yu2022surprising}. Similar to the MPPMR environment, for all our experiments in Overcooked we generate a population of $M=8$ agents for all methods per layout.

\subsection{Sample Efficiency Comparisons}
\label{efficiency_exp}

% \input{figures_and_table÷s/tab_overcooked_spxp_results}

% \input{figures_and_tables/fig_scaling}

% \input{figures_and_tables/fig3_trainingcurves}

% In this section we compare the training efficiency of our full method XPM-WM against other cross-play minimization baselines. We implement XPM-WM together with LIPO and CoMeDi on all baseline layouts of the popular Overcooked AI environment \citep{carroll2019utility}, which is 2-play fully-cooperative game in which players control cooks in a kitchen to serve onion soups.
 We plot the \textbf{total number of environment steps} for each method against the number of agents in $\mathcal{D}$ for all 5 Overcooked base layouts. We include self, cross and mixed play trajectories required for each method and show the plot in Figure \ref{fig:oc_scaling}. We also provide scaling results in terms wall clock time in the Supplementary Material.
\begin{figure*}[t!]
\centering
{\includegraphics[width=\textwidth]{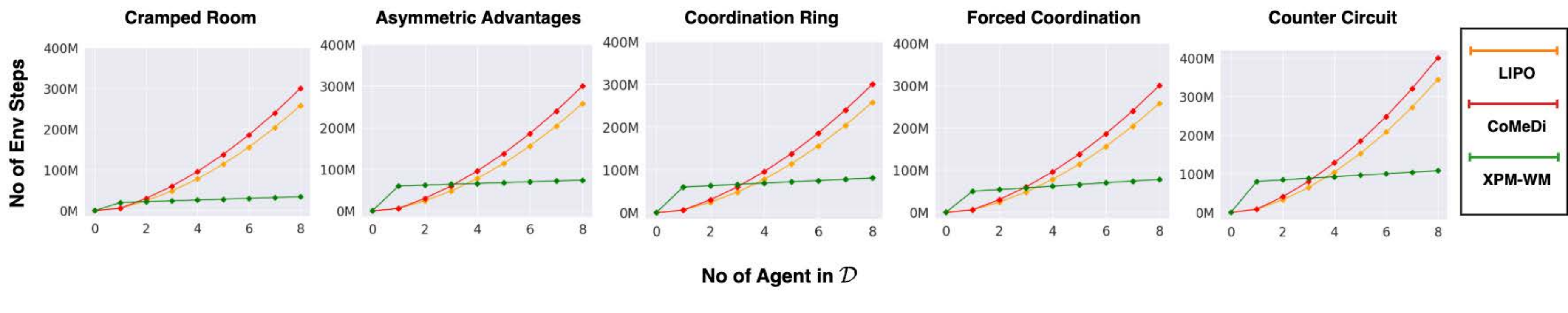}}

\vspace{-6mm}

\caption{Scaling by number of agents. We show the training cost in terms of environmental steps for all 5 Overcooked layouts as number agents $M$ in the population increases. LIPO and CoMeDi both have non-linear scaling in environmental steps required as $M$ increases due to the requirement to sample additional trajectories whereas XPM-WM has a linear scaling after the initial agent that is trained together with the world model.}
\label{fig:oc_scaling}
\vspace{-4mm}
\end{figure*}

\textbf{Results:}
We show that though training the initial agent with World Model incurs a higher training cost than normal SP, the framework of using a pre-trained world model results in an overall more sample efficient method as the number of agents, $M$ increases. We also demonstrate that the training cost in terms of environmental steps associated with XPM-WM scales linearly with $M$, whereas other XPM methods exhibit superlinear scaling with $M$. The improvement in sample efficiency comes from the significantly reduced training steps needed for subsequent agents due to the bootstrapping of the pre-trained World Model. 
% We show the comparison in terms of training steps for a single agent in the Supplementary Material \ref{oc_traiing_curves}. 
Furthermore, the use of simulated trajectories also amortizes the cost of sampling additional XP and MP trajectories. LIPO requires sampling $bm$ trajectories per iteration, where $b$ is the training batch size and $m$ the current number of agents in the population. As $m$ increases, so does the training cost per iteration. Similarly, CoMeDi requires $b(m+1)$ trajectories. As XPM-WM simulates SP and XP trajectories, we only require a constant $b$ trajectories per iteration to update the world model.

\subsection{Population Diversity Evaluation}
\label{sp_xp_score}

\begin{wraptable}{r}{0.55\textwidth}
    \centering
    \vspace{-5mm}
    \caption{The average number of dishes served in Self Play and Cross Play across all 5 Overcooked Environments for each method of XPM-based diverse agent population generation. Both self play and cross play scores are averaged over 20 episode of 400 steps.\{CR = Cramped Room, AA = Asymmetric Advantages, C Ring = Coordination Ring, FC= Forced Coordination, CC=Counter Circuit\}}
    \small{
    \begin{tabular}{c|cc|cc|cc}
        \toprule
        \multirow{2}{*}{Layout}& \multicolumn{2}{c|}{LIPO} & \multicolumn{2}{c|}{CoMeDi} & \multicolumn{2}{c}{XPM-WM (Ours)} \\
          & SP$\uparrow$&XP$\downarrow$&SP$\uparrow$& XP$\downarrow$&SP$\uparrow$& XP$\downarrow$\\
        \midrule
        CR & $8.99$  &   $2.40$ &$8.29$  &   $1.58$&  $10.23$  &   $0.43$ \\
        AA   & $9.07$   &   $1.76$&  $11.23$& $6.80$   & $18.28$  & $0.83$ \\
        C Ring  &  $7.77$  & $0.11$  & $9.45$&$2.74$ & $10.81$  &  $0.03$\\
        FC & $10.50$   & $1.99$  &  $8.84$ & $1.61$& $14.23$  &  $0.07$ \\
        CC    &  $8.47$  & $0.36$  &  $8.94$ &  $1.26$    & $8.25$  & $0.72$  \\
        \midrule
        Average   &  $8.95$  & $1.32$  &  $9.35$ &  $2.80$    & $\bm{12.36}$  & $\bm{0.42}$  \\
        \bottomrule
    \end{tabular}}
    \vspace{-5mm}
    \label{tab:oc_sp_xp}
\end{wraptable}

To demonstrate the diversity of the generated agents, we show the average of SP and XP scores of the agents and compare them with the other XPM baselines. XP scores are computed by averaging over every pair wise return with every other agent within the generated population $\mathcal{D}$ at both starting positions. We evaluate the SP and XP scores across all 5 base Overcooked layouts and show the results in Table \ref{tab:oc_sp_xp}.

\textbf{Results:}
We show that agents generated with XPM-WM has the widest SP-XP gap among all three XPM methods, which demonstrates that each agent is an effective agent (High SP) that are mutually incompatible with one another (Low XP). We also show that XPM-WM has the highest average SP scores across all 5 base Overcooked layouts, while requiring significantly less cumulative cost in terms of environment interactions (as seen in Figure \ref{fig:oc_scaling}).

% We show that XPM-WM is significantly more sample efficient in training cross-play minimization agents across all $5$ Overcooked layouts, with all layouts at least $4$ times more sample efficient than CoMeDi, which is the closest method in terms of performance. Simpler environments such as cramped room are $8$ and $6$ times more sample efficient compared to CoMeDi and LIPO respectively for achieving comparable performance in self-play. The efficiency gains also make XPM-WM a much more scalable method for training populations of larger sizes.

% \input{figures_and_tables/tab_overcooked_spxp_results}

\subsection{Ego Agent Results}
% \label{ego_exp}

% We now test if training a population with XPM-WM has comparable diversity compared to our baselines. We test this by computing the average self and cross-play returns across all $5$ Overcooked layouts. We show the average returns in Table \ref{tab:oc_sp_xp}

% Our results show that population trained with XPM-WM achieved simultaneously the highest average self-play performance and the lowest self-play performance across all $5$ Overcooked layouts, which suggests that XPM-WM not only learned very performant agents but are also meaningfully diverse from one another.

% \input{figures_and_tables/tab3_userstudyanalysis}

\subsubsection{XPM-WM}
\label{ego_exp}

% \begin{wraptable}{r}{7.3cm}
\begin{wraptable}{r}{0.57\linewidth}
    % \centering
    \vspace{-5mm}

    \caption{Results for Ego agents averaged across 5 Overcooked layouts . For each entry we compute the mean number dishes served as well as the standard error over 40 XP episodes for each partner agent in holdout population. E-X denotes an Ego agent trained with partner agent generated using X method.}
    {\footnotesize
    \begin{tabular}{c|ccc}
        \toprule
        &E-LIPO &E-CoMeDi &E-XPM-WM (Ours)\\
        \midrule
        % Self-Play  &&&&& \\
        % MEP \cite{zhao2023maximum}&&&&&\\
        LIPO     & $5.66 \pm 0.07$& $5.34 \pm 0.06$&  $4.88 \pm 0.07$\\   
        CoMeDi  &   $6.42 \pm 0.07$& $6.25 \pm 0.06$&   $6.18 \pm 0.06$\\    
        XPM-WM (Ours)&  $4.88 \pm 0.07$& $3.75 \pm 0.07$ &$6.62 \pm 0.07$ \\
        % (Ours) &&
        \midrule
        
        Average &  $5.66 \pm 0.04$& $5.11 \pm 0.04$& $\bm{5.80 \pm 0.04}$\\
        \bottomrule
    \end{tabular}}

    % \begin{tabular}{c|ccc}
    %     \toprule
    %     &E-LIPO &E-CoMeDi &E-XPM-WM \\
    %     \midrule
    %     LIPO     & $2.18$  & $2.25$&  $3.17$ \\   
    %     CoMeDi  &   $4.54$   & $3.16$  &   $5.99$  \\
    %     XPM-WM  &  $1.68$ & $1.35$ & $4.41$ \\
    %     \midrule
    %     Average &  $2.80$ & $2.25$ & $\bm{4.52}$ \\
    %     \bottomrule
    % \end{tabular}
    % \vspace{-4mm}
    
    \label{tab:oc_ego}
\vspace{-6mm}
\end{wraptable}
% \end{wraptable}

We train  ego agents with the agent population generated using LIPO, CoMeDi and XPM-WM across three random seeds. We use HiPT \citep{loo2023hierarchical} as the ego agents. We then train a separate holdout population for each baseline to validate the ego-agents' generalizability to novel partners. We present the average pair-wise rewards, in terms of number dishes served (Ego agent with hold-out populations) across all 5 Overcooked layouts in Table \ref{tab:oc_ego} and provide layout specific results in the Supplementary Material.

\textbf{Results:}
Our results show the ego agent trained with XPM-WM achieved the similar average score across the 3 different holdout populations as compared to LIPO and CoMeDi. This demonstrates that the population generated by XPM-WM is comparable in the other baseline XPM methods in terms of effectiveness as a training population for ZSC agents. Furthermore we also observe that the held-out XPM-WM agents present more of a challenge to E-LIPO and E-CoMeDi agents as compared to other held of populations in terms of the mean number dishes served.

\subsubsection{Overcooked v2}
\label{sec:oc_v2}

To evaluate the effectiveness of XPM-WM on partially observable environments, we evaluate XPM-WM on the Asymmetric Advantages with Center Recipe Layout from Overcooked v2 \citep{gesslerovercookedv2}. The layout is an augmented version of Asymmetric Advantages with observable grid of size $R$ around the player. Additionally, there is a recipe indicator at the center of the layout which indicates which recipe to serve. For our experiment, we set $R = 1$ for a $3 \times 3$ observation grid. Similar to our previous experiments, we train an ego agent generated using XPM-WM, LIPO and CoMeDi population of size $M=8$ and evaluate the resultant ego agents on held out populations from the three respective methods. We show the performance in terms of mean rewards in Table \ref{tab:ocv2_ego}.

\begin{wraptable}{r}{0.6\linewidth}
    \centering
    \vspace{-4.5mm}

    \caption{Results for XPM Ego agents on Overcooked v2 Asymmetric Advantages with center receipe layout. The evaluation framework is similar to that of Table \ref{tab:oc_ego}.}
    \footnotesize{
     \vspace{-2mm}
    \begin{tabular}{c|ccc}
        \toprule
        &E-LIPO &E-CoMeDi &E-XPM-WM (Ours)\\
        \midrule
        % Self-Play  &&&&& \\
        % MEP \cite{zhao2023maximum}&&&&&\\
        LIPO      & 13.75 $\pm$ 0.43 &12.54 $\pm$ 0.48& 15.11 $\pm$ 0.41 \\   
        CoMeDi  &   19.20 $\pm$ 0.25 & 18.69 $\pm$ 0.25 &   19.87 $\pm$ 0.24 \\
        
        XPM-WM (Ours)&  7.71 $\pm$ 0.52 & 8.58 $\pm$ 0.51 & 10.17 $\pm$ 0.52\\
        \midrule
        Average &   13.55 $\pm$ 0.28 & 13.27 $\pm$ 0.28 & $\bm{15.05 \pm 0.27}$ \\
        \bottomrule
    \end{tabular}}
     \vspace{-4mm}
    
    \label{tab:ocv2_ego}
\end{wraptable}

\textbf{Results:} 
Similar to results for the fully-observable Overcooked layouts, we found that the ego agent trained with XPM-WM  has the highest score across the three held out evaluation populations. We also observe a similar phenomena whereby E-LIPO and E-CoMeDi shows a significant drop in performance when evaluated on XPM-WM agents.

\subsubsection{FCP-WM and MEP-WM}
\label{fcp_score}

Similar to XPM-WM, we train ego agents with populations of size $M=8$ generated using FCP-WM and MEP-WM and evaluate the ego agents on held out sets of FCP(SP) and MEP respectively. We compare the performance to the ego agent trained with the base FCP and MEP algorithms and evaluate them with held out populations from FCP and MEP on all five base Overcooked layouts. For FCP's held out population we only evaluate on fully trained agent, i.e. the final checkpoint. To ensure a fair comparison, we also train the base MEP method in a sequential manner as described in Section {\ref{sec_mep_fcp}}. For both MEP and MEP-WM, we set the entropy reward coefficient, $\alpha$ to $0.01$ for all layouts except Counter Circuit where it is set to $0.005$. We show the results in Table \ref{tab:oc_fcp_mep}.

% \begin{table}[H]
% \vspace{-2mm}

%     \centering
%     \caption{Results of Ego agents trained with FCP-WM and FCP evaluated on a held out Self-Play population. Similar to \ref{tab:oc_ego} we compute the mean dishes served over 40 episodes as well as standard error.}
%     \vspace{-2mm}

%     \begin{tabular}{c|cc}
%         \toprule
%         & E-FCP-WM & E-FCP \\
%         \midrule
%         Cramped Room & 10.22 $\pm$ 0.11 & 7.47 $\pm$ 0.11\\
%         Asymmetric Advantages   & 8.31 $\pm$ 0.35& 4.88 $\pm$ 0.28 \\
%         Coordination Ring  &  6.86 $\pm$ 0.09 & 5.66 $\pm$ 0.10 \\
%         Forced Coordination & 4.70 $\pm$ 0.26 & 7.88 $\pm$ 0.15\\

%         \midrule
%         Average   & $\bm{7.52 \pm 0.13}$ & 6.47 $\pm$ 0.09\\
%         \bottomrule
%     \end{tabular}
    
%     \label{tab:fcp}
% \vspace{-2mm}

% \end{table}

\begin{wraptable}{r}{0.6\textwidth}
% \parbox{.48\linewidth}{
\vspace{-4mm}

\centering
\caption{Results of Ego agents trained with FCP-WM, MEP-WM evaluated on a held out Self-Play population generated using FCP and MEP respectively. Similar to \ref{tab:oc_ego} we compute the mean dishes served over 40 episodes as well as standard error.}
\small{
\begin{tabular}{c|cc|cc}
        \toprule
        & E-FCP-WM & E-FCP & E-MEP-WM & E-MEP\\
        \midrule
        CR & 7.38 $\pm$ 0.15 & 7.47 $\pm$ 0.11  & 6.63 $\pm$ 0.22  &5.70 $\pm$ 0.21 \\
        AA   & 8.41 $\pm$ 0.35& 4.87 $\pm$ 0.28 &  4.10 $\pm$ 0.2& 8.15 $\pm$ 0.29 \\
        C. Ring  &  6.82 $\pm$ 0.09 & 5.72 $\pm$ 0.11 &   3.29 $\pm$ 0.13  & 5.70 $\pm$ 0.18 \\
        FC & 4.73 $\pm$ 0.26 & 8.00 $\pm$ 0.15 & 2.32 $\pm$ 0.17 & 6.13 $\pm$ 0.35 \\
        CC & 3.54 $\pm$ 0.11 & 3.77 $\pm$ 0.09& 1.79 $\pm$ 0.16 & 5.57 $\pm$ 0.16\\

        \midrule
        Average   & $\bm{6.18 \pm 0.11}$ & 5.97 $\pm$ 0.08& 3.38 $\pm$ 0.10 & $\bm{6.25 \pm 0.11}$\\
        \bottomrule
    \end{tabular}
    \vspace{-3mm}
}
\label{tab:oc_fcp_mep}
% \hfill
% \parbox{.48\linewidth}{
% \centering
% \caption{Results of Ego agents trained with MEP-WM and MEP evaluated on a held out MEP population. The evaluation is similar to that of Table \ref{tab:oc_ego}}
% \small{
% \begin{tabular}{c|cc}
%         \toprule
%         & E-MEP-WM & E-MEP \\
%         \midrule
%         CR & 2.47 $\pm$ 0.20  &5.70 $\pm$ 0.29 \\
%         AA   &  5.66 $\pm$ 0.37& 8.38 $\pm$ 0.41 \\
%         C. Ring  &   0.36 $\pm$ 0.07  & 5.61 $\pm$ 0.26 \\
%         FC & 4.74 $\pm$ 0.36 & 6.27 $\pm$ 0.50 \\

%         \midrule
%         Average   & 3.31 $\pm$ 0.16 &  6.49 $\pm$ 0.19\\
%         \bottomrule

% \end{tabular}
% }}
\end{wraptable}

\textbf{Results:}
We show the ego agent trained with FCP-WM, E-FCP-WM on average matches the ego agent trained FCP, E-FCP. Which demonstrates that the WM training paradigm can be adapted to FCP. However the ego agent trained with MEP-WM significantly under-performs the ego agent trained using MEP. We hypothesize that this could be due to the compounding error of predicting both the task and entropy rewards using imaginary trajectories. The log probability of the average policy requires accurate log probabilities of every prior agent at each simulated state. Hence, any inaccuracies from a single agent can potentiality compound the inaccuracy in the entropy reward, whereas for XPM-WM and FCP, this is less of an issue as agents directly act on the simulated states and the policies are only trained using the simulated rewards.

% \subsubsection{XPM-WM on Overcooked v2}
% \label{sec:oc_v2}

% \input{figures_and_tables/tab_overcookedv2_ego_results}

% To evaluate the effectiveness of XPM-WM on partially observable environments, we evaluate XPM-WM on the Asymmetric Advantages with Center Recipe Layout from Overcooked v2 \citep{gesslerovercookedv2}. The layout is an augmented version of Asymmetric Advantages with observable grid of size $R$ around the player. Additionally, there is a recipe indicator at the center of the layout which indicates which recipe to serve. For our experiment, we set $R = 1$ for a $3 \times 3$ observation grid. Similar to our previous experiments, we train an ego agent generated using XPM-WM, LIPO and CoMeDi population of size $M=8$ and evaluate the resultant ego agents on held out populations from the three respective methods. We show the results in Table \ref{tab:ocv2_ego}.

% % \input{figures_and_tables/fig_user_studies_bar}

% \textbf{Results:} 
% Similar to results for the fully-observable Overcooked layouts, we found that the ego agent trained with XPM-WM  has the highest score across the three held out evaluation populations. We also observe a similar phenomena whereby E-LIPO and E-CoMeDi shows a significant drop in performance when evaluated on XPM-WM agents.

\subsection{User Studies }
\label{user_studies}
\begin{figure*}[t!]
  \vspace{-6mm}
  \centering
  \includegraphics[width=\linewidth]{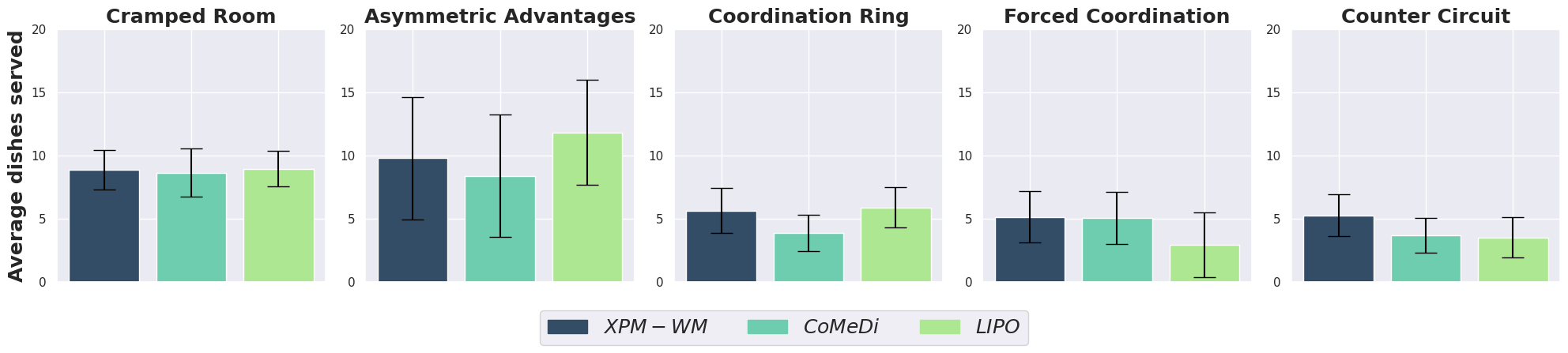}
  \caption{A comparison of the user study scores for HiPT ego agent trained with XPM-WM, CoMeDi and LIPO. Error bars represent the standard deviation.}
  \label{fig:user_studies_rewards}
  \vspace{-6mm}
\end{figure*}

We then conduct an IRB approved user study by pairing 40 real human partners with each of the 3 ego agents used in Section \ref{ego_exp}. Each human partner plays a total of 20 games with randomized layout and ego agent partners. We compare the average scores of each of the 3 ego agents when partnered with human partners and visualize the results in Figure \ref{fig:user_studies_rewards}.

\textbf{Results:}
We find that the ego agents trained with population generated via XPM-WM achieves similar scores as compared to other ego agents, which suggest a comparable population diversity in XPM-WM as compared to other baselines. We have also found the results in the user studies to be consistent with the results on the hold out partner population.

\subsubsection{Diversity Analysis}
\label{diversity_analysis}

We then conduct further qualitative analysis on whether the ego agents can adapt to the diverse conventions present in the human partners. To do so, we isolate three common behaviors: plating frequency, onion picking frequency, and amount of movement in the environment. We then determine if a human partner has a high or low preference towards the behavior if their frequency/amount of said behavior is half standard deviation away from the mean of all humans for that agent. We then compute the average score for partners of said convention and show how much it deviates from the score with all human conventions. We show this comparison in Table \ref{tab:preferences_comparison}. Overall our analysis shows that the ego agent trained with the XPM-WM generated population exhibits the least amount of deviation across all 6 conventions as compared to the other two baselines, suggesting the population is sufficiently diverse to cover a variety of human behavioral preferences.

\begin{table}[H]
    \centering
    \vspace{-3mm}
    \caption{Comparison of difference in average number of dishes served with specific human conventions relative to all humans across all Overcooked environments.}
    \footnotesize{
    \begin{tabular}{c|cc|cc|cc}
        \toprule
        \multirow{2}{*}{} & \multicolumn{2}{c|}{Plating Preference} & \multicolumn{2}{c|}{Onion Picking Preference} & \multicolumn{2}{c}{Movement Activity Preference} \\
        & High  & Low & High& Low& High & Low  \\
        \midrule
        E-XPM-WM (Ours) & $0.74$ & $-0.43$ & $ 1.3$ & $-1.33$ & $0.17$ & $-0.20$ \\
        E-LIPO          & $1.76$  & $-1.17$ & $1.14$ & $-0.4$ & $1.13$ & $-0.51$ \\
        E-CoMeDi        & $1.72$ & $-1.87$  & $0.90$ & $-0.44$ & $0.30$  & $-0.59$ \\
        \bottomrule
    \end{tabular}}
    % \caption{Comparison of difference in average number of dishes served with specific human conventions relative to all humans across all environments.}
    \label{tab:preferences_comparison}
\end{table}

\section{Conclusion}
In this work, we introduced XPM-WM, an efficient method for generating diverse agents for Zero-Shot Coordination. Our method is able to train agents with cross-play minimization in a more efficient manner by first learning a World Model and then learning via simulated trajectories. We show that our method is significantly more sample efficient compared to the current XPM methods when training individual agents while achieving the same level of performance in terms of population diversity. We also show our method is more scalable in terms of cumulative training cost as compared to the other methods when the number of agents increase. Finally, we show the World Model training framework is also extendable to other population training methods beyond XPM.

% \hl{
\textbf{Limitations:}
Though the World Model training paradigm significantly speeds up during population training in-terms of environmental steps, the method still largely hinges on a well-trained world model that can accurately model the environment. Hence the World Model may steer subsequent agents towards states where it is "confident" in (i.e. accurately predict the transition and reward) as compared to non-WM PBT methods where agents are free to learn their own state representation from scratch. We believe however that this will become less of issue as the number of agents increase (given our easily scalable approach) and the World Model becomes robust enough to cover most of the state transitions. 
% }

\section*{Acknowledgments}
This project is jointly supported by the National Research Foundation Singapore and DSO National Laboratories under the AI Singapore Programme  (Award Number: AISG2-RP-2020-016) and the Google South \& South East Asia Research Award (2024).

% \clearpage
\bibliography{main}
\bibliographystyle{tmlr}

\clearpage
\appendix

% \end{proof}
\section{Algorithm details for XPM-WM}
\begin{algorithm}[H]
    \caption{XPM-WM Training for $m$-th Agent}
    \label{alg:mth_agent_wm}
    \begin{flushleft}
    \textbf{Input}: Set of $(m-1)$ agent partners $\mathcal{D}_{m-1} = \{\pi_{1},...,\pi_{m-1}\}$, trained World Model, $\mathcal{W}_\theta$
    \end{flushleft}
    % \textbf{Parameter}: Execution bounds for low-level policy $[p_{min},p_{max}]$, Influence reward coefficient $\gamma$, Environment reward coefficient  $\alpha$
    % \textbf{Output}: Your algorithm's output
    \begin{algorithmic}[1] %[1] enables line numbers
        % \STATE Let $t=0$.
        % \WHILE{not converged }
        \STATE Initialize actor $\bm{\pi}^{\phi_m}_m$ and critics $\{v^m_{\psi_1},...,v^m_{\psi_{m-1}},v^m_{\psi_m}\}$ for agent $m$
        \STATE Initialize self-play and cross-play replay buffer, $\mathcal{B}_{SP}$ and $\mathcal{B}_{XP}$ 
        \STATE Populate $\mathcal{B}_{SP}$ and $\mathcal{B}_{XP}$ with $K$ random episodes
        \WHILE{not converged }
        \FOR {$c=0$ {\bfseries to} $C$ training steps}
        \STATE Sample self-play and cross-play trajectories, $\bm{\tau}_{SP}$ and $\bm{\tau}_{XP}$ from $\mathcal{B}_{SP}$ and $\mathcal{B}_{XP}$
        \STATE Fine-tune World Model parameters, $\theta$ with $\bm{\tau}_{SP}$ and $\bm{\tau}_{XP}$ using Loss Function in (7)
        \STATE Simulate joint self-play trajectories, $\bm{\hat{\tau}}^{SP}_{\mathcal{W}}$ from states sampled from  $\mathcal{B}_{SP}$ and $\mathcal{B}_{XP}$ with $\bm{\pi}_{SP}$
        \STATE Determine current best cross-play partner $\bm{\pi}_{XP} = max_{\bm{\pi}_{k} \in \mathcal{D}_{m-1}}J_{XP}(\bm{\pi}_{m},\bm{\pi}_{k})$ from trajectories in $\mathcal{B}_{XP}$ 
        \STATE Simulate joint cross-play  trajectories, $\bm{\hat{\tau}}^{XP}_{\mathcal{W}}$ from  states in $\mathcal{B}_{XP}$ with $\bm{\pi}_{XP}$
        \STATE Estimate self-play returns and compute self-play objective with $\bm{\hat{\tau}}^{SP}_{\mathcal{W}}$
        \STATE Estimate cross-play returns and compute cross-play objective with $\bm{\hat{\tau}}^{SP}_{\mathcal{W}}$ 
        \STATE Compute XPM-Sim as in (3)
        \STATE Update actor parameters $\phi_m$
        \STATE Update critic parameters ${\psi_1,...,\psi_m}$
        \ENDFOR
        \STATE Collect self-play episodes from environment with $\bm{\pi}_m$ and $\mathcal{W}_\theta$.
        \STATE Add self-play episodes to $\mathcal{B}_{SP}$
        \STATE Sample random cross-play partner from $\bm{\pi}_k \sim\mathcal{D}_{m-1}$
        \STATE Collect cross-play episodes from environment with $\mathcal{W}_\theta$, $\bm{\pi}_m$ and, $\bm{\pi}_k$.
        \STATE Add cross-play episodes to $\mathcal{B}_{XP}$.
        \ENDWHILE
        % \STATE \textbf{return} $\pi_{\theta_{high}}(z|s)$, $\pi_{\theta_{low}}(a|s,z)$
    \end{algorithmic}
    \begin{flushleft}
    \textbf{Output}: $m$-th agent,$\bm{\pi}_m$, fine-tuned World Model, $\mathcal{W}_{\theta^m}$
    \end{flushleft}
\end{algorithm}
% \section{Algorithm details for XPM-WM}
We present the full algorithm for training an agent using XPM-WM in Algorithm \ref{alg:mth_agent_wm}. For the first agent, the training objective is solely  the SP objective without sampling any XP trajectories.

\section{Environment description}

\subsection{Multi Particle Point Mass Rendezvous \cite{charakorn2023generating}}
\begin{figure}[H]
  \centering
  \fbox{\includegraphics[width=0.25\linewidth]{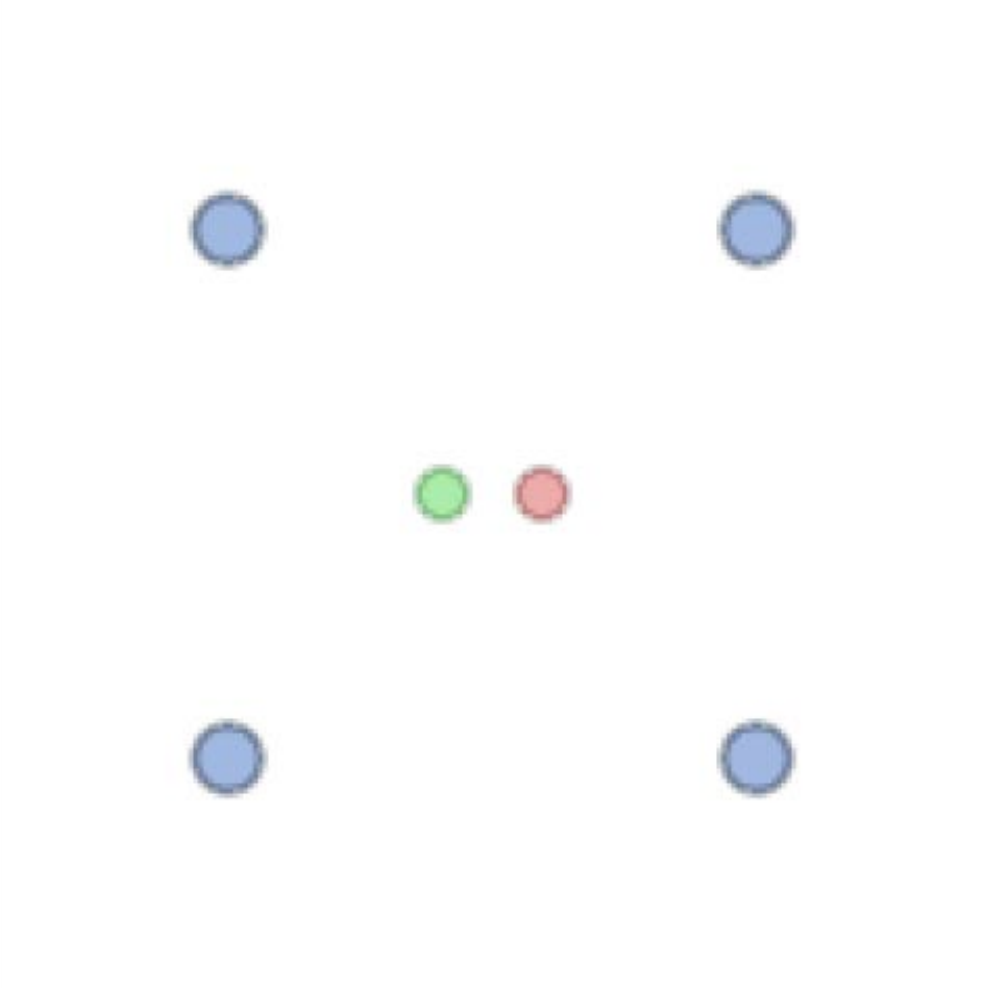}}
  \caption{The initial state of Multi Particle Point Mass Rendezvous Environment. The 2 players control the red and green particles respectively. The four blue landmarks are spaced evenly in the environment. Both players receive a high reward if both particles are near one-another and to one of the landmarks. The environment episode terminates if one of the players go out of bounds of the 3x3 space.}
  \label{fig:mpe}
  % \vspace{-9pt}
\end{figure}

In this environment, two players control separate particles that can move up, down, left, right or stay in place. Four landmarks are spaces evenly in the movable space. The goal for both players is to rendezvous at one of the four landmarks and stay there. The four landmarks represent four distinct conventions that agents pairs can converge to. Players are free to move around in the 3 by 3 unit space but the episode terminates when any player moves out-of bounds. A visualization for the initial positions of the players and landmarks can be seen in Figure \ref{fig:mpe}. The Joint Rewards at each timestep is computed by the following expression:

\eq{

    r(\mathbf{a}_t|s_t) = 1 - D(\text{nearest landmark}, \text{centroid of players pos.}) - D(\text{pos. of player1}, \text{pos. of player 2}) 

}

where $D(.,.)$ is the 2D Euclidean distance between two points. For the toy experiment in Section 4 we deterministically compute the forward dynamics of the environment using the linear dynamics provided in Multi-Particle Environment source code\footnote{https://github.com/Farama-Foundation/PettingZoo} using three scalar parameters $\{timestep, damping, sensitivity\}$.

\subsection{Overcooked AI \cite{carroll2019utility}}

In this environment, players each control two chefs in a kitchen layout. The task is deliver as many onion soups to the counter tile in $400$ time steps. The action space for each player is move left, right, up or down, no-op or to interact with an object or tile in front of them. To serve a dish of soup, players have to deposit $3$ onions in the pot, wait for the pot to finish cooking and use a bowl to collect the soup before delivering it to the serving tile. For Overcooked v2 \cite{gesslerovercookedv2}, agents can only observe a grid that of radius $R$ around the agent, resulting in a $2R+1 \times 2R+1$ viewing ggrid. 

\begin{figure*}[ht]
  \centering
  \includegraphics[width=0.18\linewidth,height=0.1\textheight]{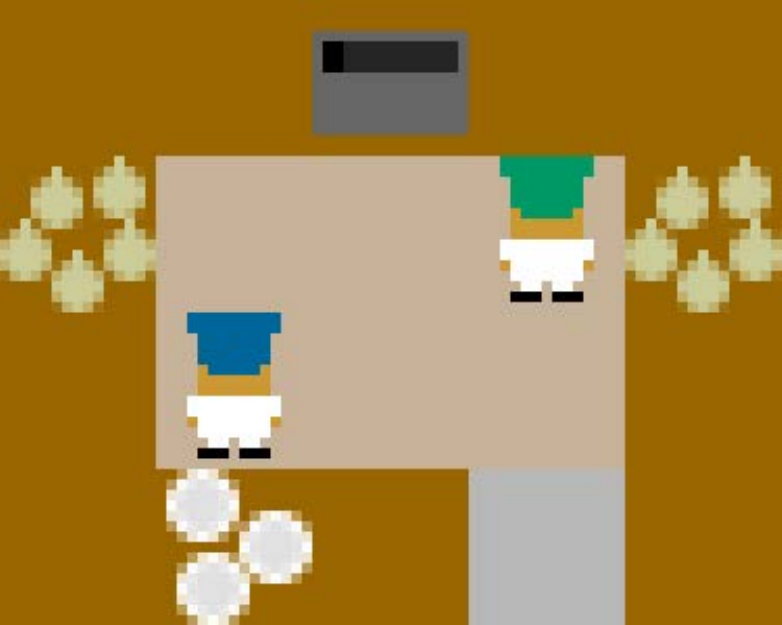}
  \includegraphics[width=0.22\linewidth,height=0.1\textheight]{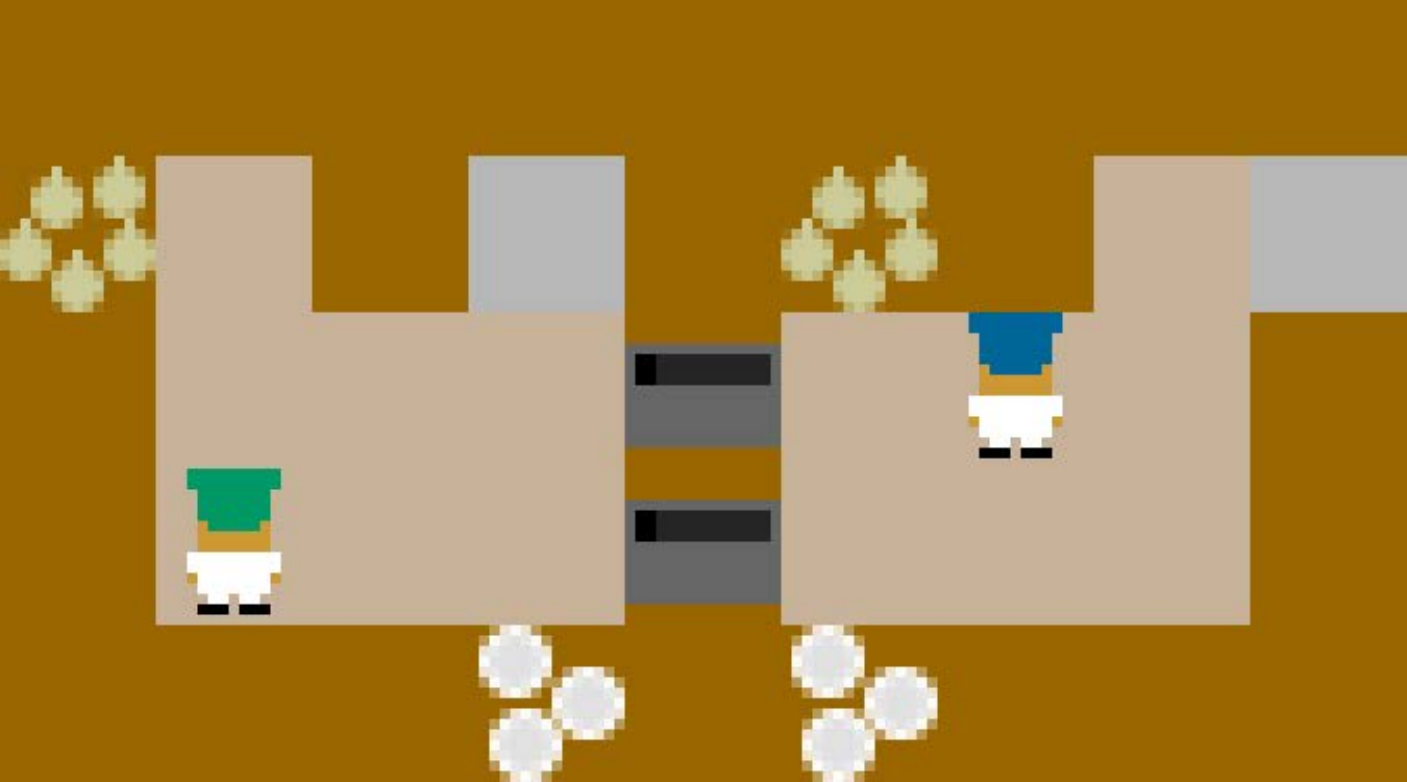}
  \includegraphics[width=0.16\linewidth,height=0.1\textheight]{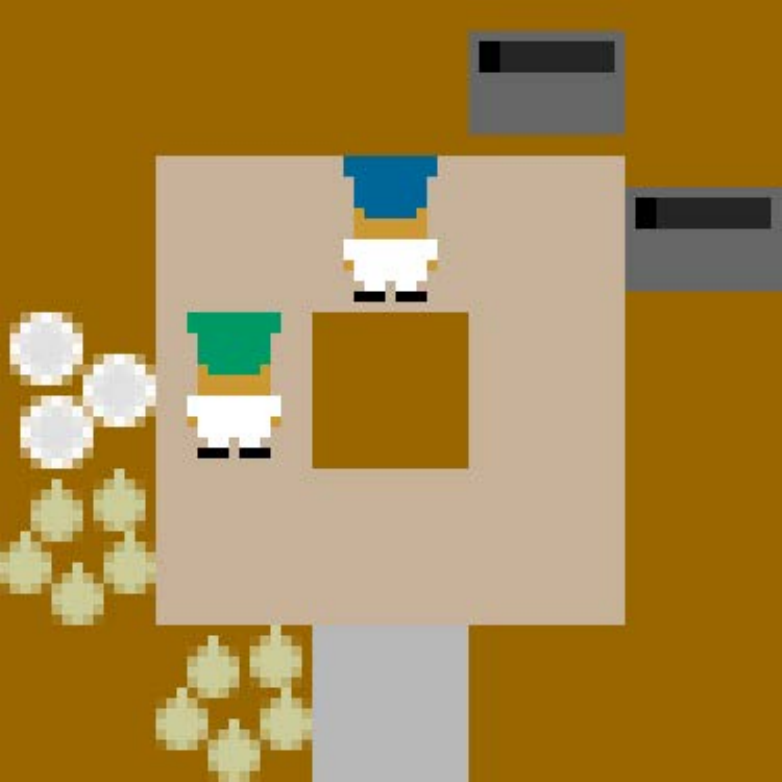}
  \includegraphics[width=0.16\linewidth,height=0.1\textheight]{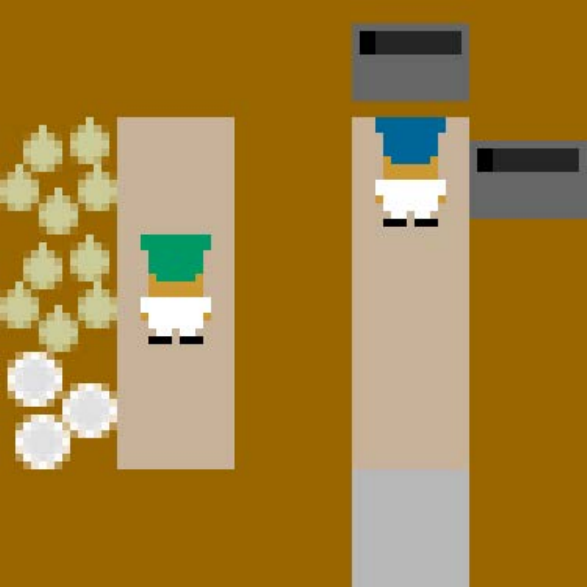}
  \includegraphics[width=0.22\linewidth,height=0.1\textheight]{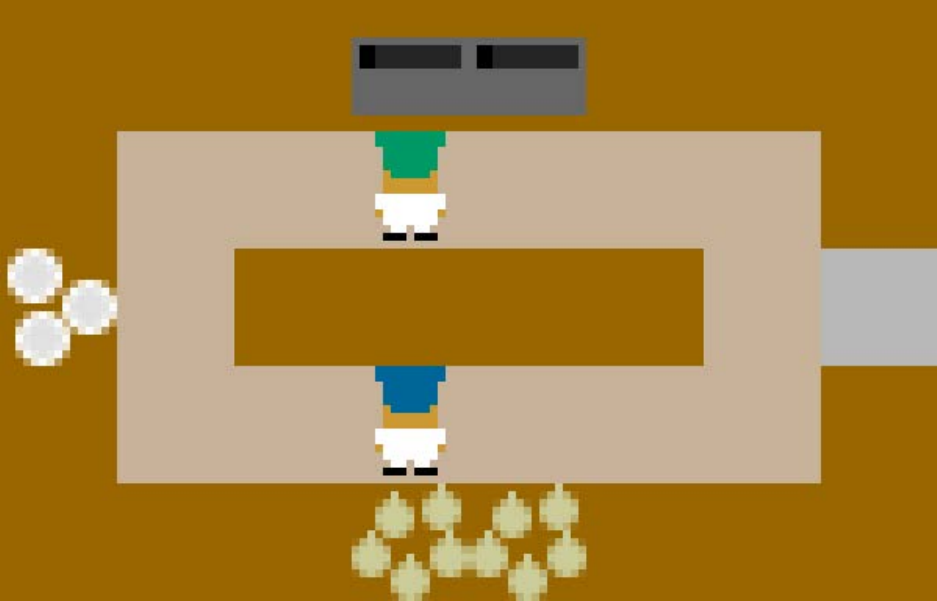}
  \includegraphics[width=0.4\linewidth,height=0.2\textheight]{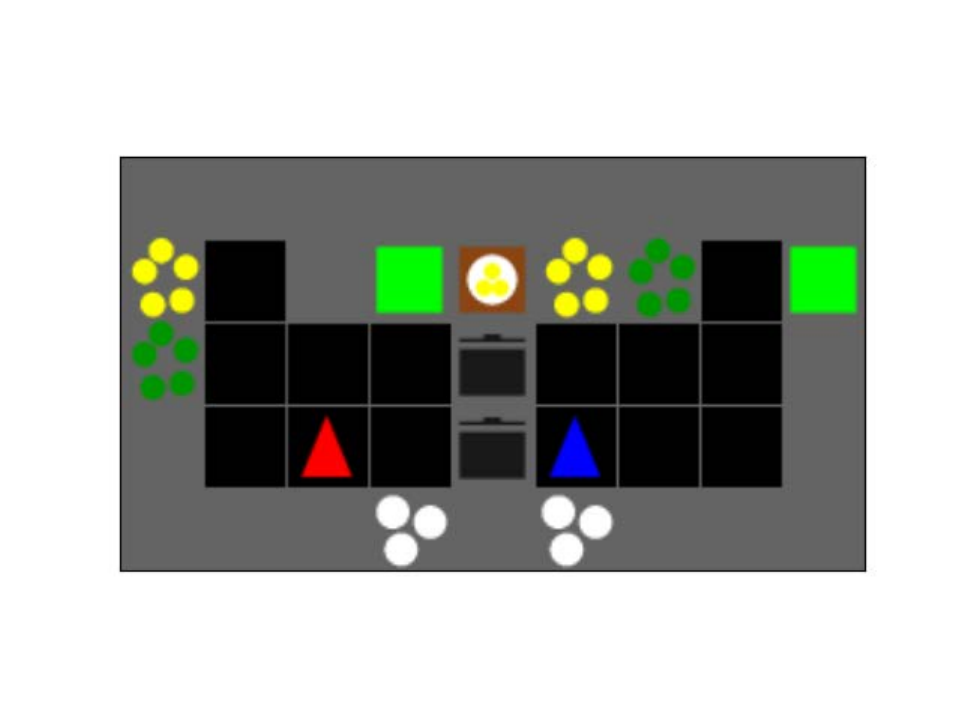}
  \caption{\textbf{The Five original Overcooked layouts} Top row from left to right: Cramped Room, Asymmetric Advantages, Coordination Ring, Forced Coordination and Counter Circuit. The Blue and Green hat chefs indicate the 2 different starting positions for each layout. \textbf{Bottom row} Asymmetric Advantages Recipes Center layout from Overcooked v2 The bowl at the center incicates which recipe to serve.}
  \label{fig:oc_layout}
\end{figure*}

We conduct our experiments on the $5$ standard layouts as wll as the partially obsevable layout from Overcooked v2. The layouts are visualized in Figure \ref{fig:oc_layout}.

% \clearpage

\section{Implementation  Details }
\label{details}

\subsection{Training Algorithms}

For the XPM-Sim agents in Section 4, we train the agents using the REINFORCE algorithm with accentralized critic and Generalized Advantage Estimation (GAE) \cite{schulman2015high} for advantage estimation.  For the XPM-WM, XPm-FCP and XPM-MEP Agents in  Section 5, We adopt the Dreamer objective and train agents using REINFORCE  with the recursive $\lambda$-target \cite{hafnermastering} for target returns estimation. For MEP, FCP, LIPO and CoMeDI, we train all agents using MAPPO \cite{yu2022surprising} with GAE. 

\subsection{Vectorized Reward Prediction}
\label{vec_rew_details}

As described in Section 5,  the XPM-WM reward predictor outputs a binary vector reward based on events which are then scaled by constant value. Here we describe the specific events for Overcooked AI as well as Overcooked v2 and the scaling factor for each event. Note that events are player specific but the final joint reward is identical for both players.

\begin{table}[H]
    \centering
    \caption{Reward Events for the Overcooked AI environment}
    \begin{tabular}{cc}
        \toprule
        Event & Reward scaling \\
        \midrule
          Player $i$ picks up ingredient &  $1$\\
          Player  $i$ picks up a serving bowl & $1$ \\
          Player $i$ puts ingredient in pot & $1$\\
          Player $i$ picks up soup from pot & $3$\\
          Player $i$ delivers soup &$12$\\
        \bottomrule
    \end{tabular}
    
    \label{tab:rew_event}
\end{table}

\begin{table}[H]
    \centering
    \caption{Reward Events for the Overcooked v2 environment}
    \begin{tabular}{cc}
        \toprule
        Event & Reward scaling \\
        \midrule
          Player  $i$ picks up a serving bowl & $3$ \\
          Player $i$ puts ingredient in pot & $3$\\
          Player $i$ picks up soup from pot & $5$\\
          Soup starts cooking in pot & $5$ \\
          Player $i$ delivers soup &$20$ \\
        \bottomrule
    \end{tabular}
    
    \label{tab:rew_event_v2}
\end{table}

\subsection{Hyperparameters}

\begin{table}[H]
    \centering
    \caption{Hyperparameters for MPPMR Agents}
    \begin{tabular}{cc}
        \toprule
        hyperparameter & value \\
        \midrule
        No. Actor MLP Layers & 5 \\
        No. Critic MLP Layers & 5 \\
        MLP Layer Size & 512 \\ 
        activation & LeakyRELU \\
        LIPO $\lambda_{XP}$ & 0.25 \\
        CoMeDi $\lambda_{XP}$ &0.5 \\
        CoMeDi $\lambda_{MP}$ &0.25 \\
        XPM-Sim $\lambda_{XP}$ &0.5 \\
        SP, XP and MP Training steps & 8e5 \\
        Simulation Horizon (XPM-Sim) & 30 \\
        learning rate & 5e-4 \\
        Buffer size (XPM-Sim only) & 2e5 \\
        entropy coef. & 1e-4 \\

        \bottomrule
    \end{tabular}
    
    \label{tab:hp_mpe}
\end{table}

\begin{table*}[t!]
    \centering
    \caption{Hyperparameters for MAPPO (FCP, MEP, LIPO and CoMeDi) agents on Overcooked AI. XP coefficients, Training Steps, learning rate and Entropy coefficients are listed in order of layouts as in Figure \ref{fig:oc_layout}.}
    \begin{tabular}{cc}
        \toprule
        hyperparameter & value \\
        \midrule
        CNN Layers & 3 \\
        CNN filter size & 5x5, 3x3, 3x3 \\
        MLP Layers & 2 \\
        No. Actor MLP Layers & 1 \\
        No. Critic MLP Layers & 1 \\
        MLP Layer Size & 512 \\ 
        activation & LeakyRELU \\
        LIPO $\lambda_{XP}$ & 0.25, 0.25,0.25,0.1,0.1,0.1 \\
        CoMeDi $\lambda_{XP}$ &0.25, 0.25,0.25,0.1,0.1,0.1\\
        CoMeDi $\lambda_{MP}$ &1 \\
        SP, XP and MP Training steps (per partner)& 6e6, 6e6, 6e6, 6e6, 6e6, 8e6 \\
        learning rate & 6e-4,1e-3,8e-4,6e-4, 6e-4,1e-3 \\
        entropy coef. & 3e-3,4e-3,5e-3,8e-3,8e-3, 4e-3 \\
        \midrule
        Overcooked v2 Specific Parameters &\\
        \midrule
        Joint Critic CNN Layers & 3 \\
        Actor CNN Layers & 3 \\
        Actor RNN (GRU) layers & 1 \\

        \bottomrule
    \end{tabular}
    
    \label{tab:hp_oc}
\end{table*}

\begin{table*}[t!]
    \centering
    \caption{Hyperparameters for XPM-WM on Overcooked AI Layout. Training steps are listed in order of  layouts as in Figure \ref{fig:oc_layout}.}
    \begin{tabular}{cc}
        \toprule
        hyperparameter & value \\
        \midrule
        World Model Specific Parameters &\\
        \midrule
        Encoder & 3 x CNN, 3 x MLP\\
        Decoder & 3 x MLP, 3 x Transpose CNN \\
        Transition Predictor & 1 x GRU\\
        Reward Predictor & 5 x MLP\\
        Continue Predictor & 5 x MLP\\
        Prior & 2 x MLP \\
        Posterior & 2 x MLP \\
        CNN filter size & 5x5, 3x3, 3x3 \\
        GRU \& MLP Layer Size & 512 \\
        Joint latent dimensions & 28 \\
        Agent specific latent dimensions & 4 \\
        Latent classes & 32 \\
        Update interval & 50 \\
        WM Learning rate & 2e-4 \\
        \midrule
        Agent Specific Parameters &\\
        \midrule
        Buffer Size & 1e6 \\
        Simulation Horizon & 15 \\
        No. Actor MLP Layers & 5 \\
        No. Critic MLP Layers & 5 \\
        MLP Layer Size & 512 \\ 
        activation & LeakyRELU \\
        Phase 1  Training steps & 2e7, 6e7, 6e7, 5e7, 8e7, 8e7 \\
        Phase 2 Total Training steps per agent (SP+XP) & 2e6, 2e6, 3e6, 4e6, 4e6, 6e6 \\
        XPM-Sim $\lambda_{XP}$ &0.25 (Other layouts), 0.1(Counter Circuit) \\
        actor learning rate & 4e-5\\
        critic learning rate & 1e-4\\
        entropy coef. & 0.03 \\

        \bottomrule
    \end{tabular}
    
    \label{tab:hp_xpmwm}
\end{table*}

\begin{table}[H]
    \centering
    \caption{Hyperparameters for HiPT Ego Agent.}
    \begin{tabular}{cc}
        \toprule
        hyperparameter & value \\
        \midrule
        CNN Layers & 3 \\
        CNN filter size & 5x5, 3x3, 3x3 \\
        MLP Layers & 2 \\
        LSTM Layers & 1 \\
        No. Actor MLP Layers & 1 \\
        No. High Level Actor MLP Layers & 1 \\
        No. Critic MLP Layers & 1 \\
        LSTM \& MLP Layer Size & 512 \\ 
        activation & LeakyRELU \\
        Training steps & 1e8 \\
        learning rate & 1e-3\\
        entropy coef. & 0.01 \\

        \bottomrule
    \end{tabular}
    
    \label{tab:hp_hipt}
\end{table}

\section{Additional User Study Details}

\begin{figure*}[ht!]
  \centering
  \includegraphics[width=0.75\linewidth]{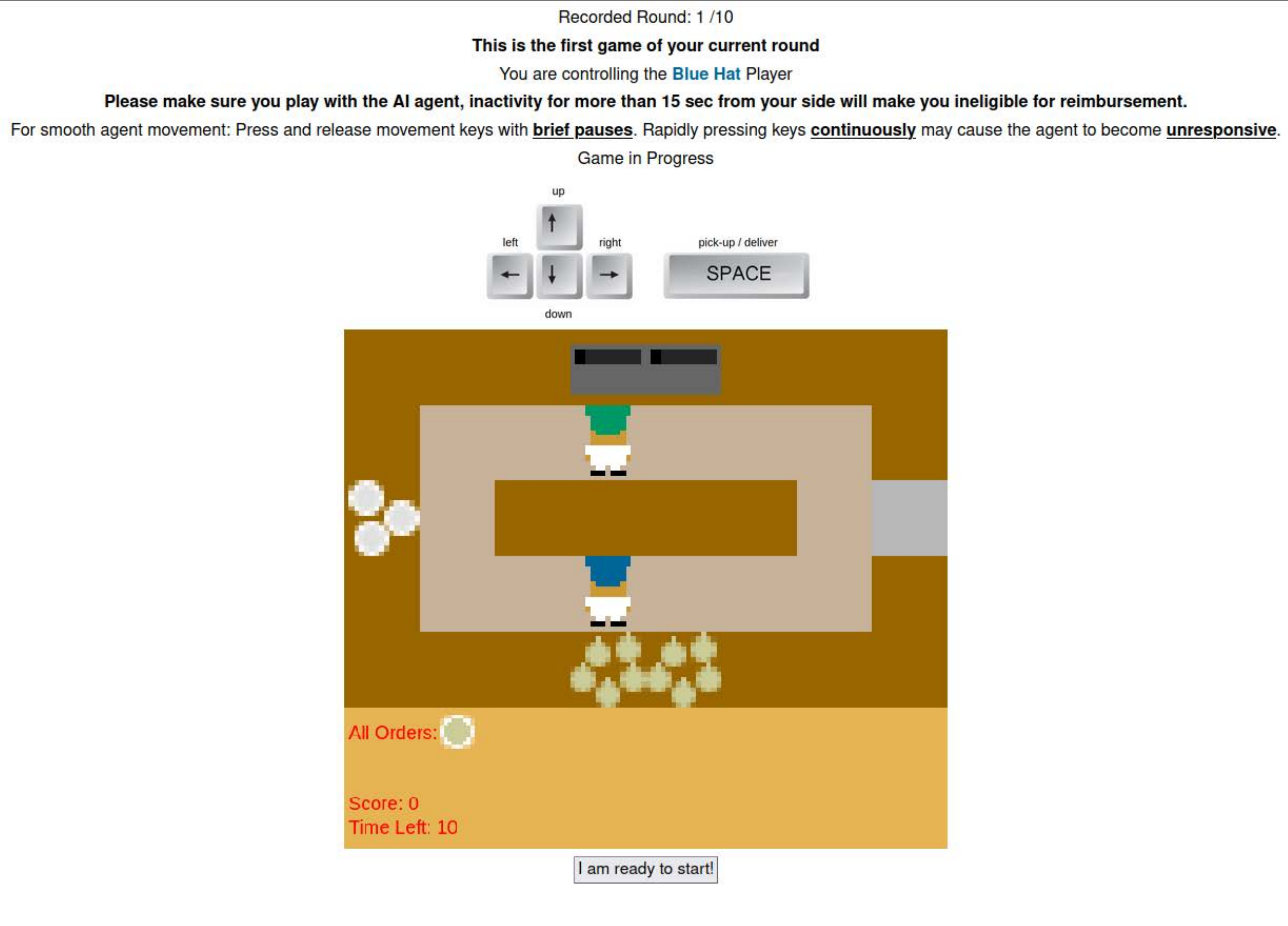}
  \caption{Screenshot of an ongoing round of game in our user studies}
  \label{fig:user_study_sc}
  % \vspace{-9pt}
\end{figure*}

\label{user_details}
% All the human partners were recruited from our local university population (N=40, Woman=20, Man=20, Median Age Range=18-25 years old) via an email invitation. Each participant plays a total of 20 games. Each games lasts lasts T=400 steps (one minute). At the start of the study, each participant is instructed clearly on the rules of the Overcooked AI environment as well as the number of games they are expected to complete. At the end of the study, each participant is asked to fill in an online form that reveals their age, gender and experience with video games. Each participant is then compensated monetarily. 
The human-AI interaction portion of this research was approved by our institutions IRB.

All the human partners were recruited from our local university population (N=40, Woman=20, Man=20, Median Age Range=18-25 years old) via an email invitation to our in-person user study and reimbursed monetary compensation for approximately 30 minutes of their time. We include screenshots from our user study in Figure \ref{fig:user_study_sc}.

\section{Qualitative Results for MPPMR}

Here we show qualitative results for the MPPMR Toy experiment. We plot the trajectories of two agents in $\mathcal{D}$ in the MPPMR environment during cross-play from all three XPM methods. We show the agents trained under LIPO learns to go out-of-bounds while agents trained with CoMeDi and XPM-Sim stay within bounds and converge to separate landmarks. 

\begin{figure}[H]
  \centering
  \subfigure[LIPO]{\fbox{\includegraphics[width=0.24\linewidth]{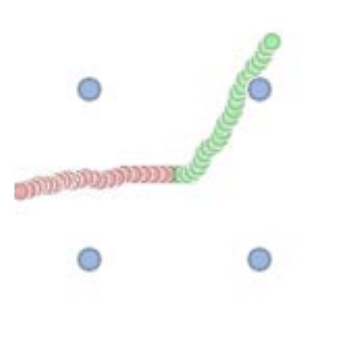}} \label{fig:human_preference_a}}
  \subfigure[CoMeDi]{\fbox{\includegraphics[width=0.24\linewidth]{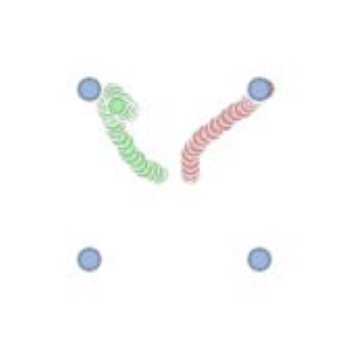}\label{fig:mpe_comedi}}}
  \subfigure[XPM-Sim]{\fbox{\includegraphics[width=0.24\linewidth]{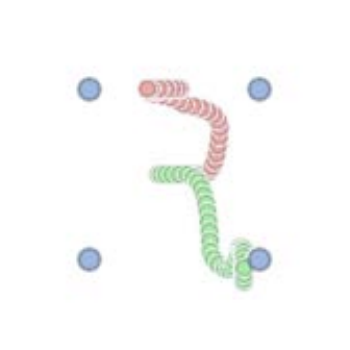}\label{fig:coedi_imag_mpe}}}

  \caption{Trajectories of 2 sample agents under cross-play in the  MPPMR environment.}
  \label{fig:mpe_qualitative}
\end{figure}

\section{Additional Results For Overcooked AI}

In this section we show the SP training curves of XPM-WM as compared to other baseline methods. We plot the training curves of each subsequent agent, i.e. agents m=2 to m=8. Our results, which we show in Figure \label{oc_training_curves} shows the increase in sample efficiency and performance of an agent that is bootstrapped by a pre-trained World Model versus a agent that is trained from scratch.

% \clearpage
\subsection{Training Curves}
\label{oc_traiing_curves}

\begin{figure}[H]
\centering
\includegraphics[width=\textwidth]{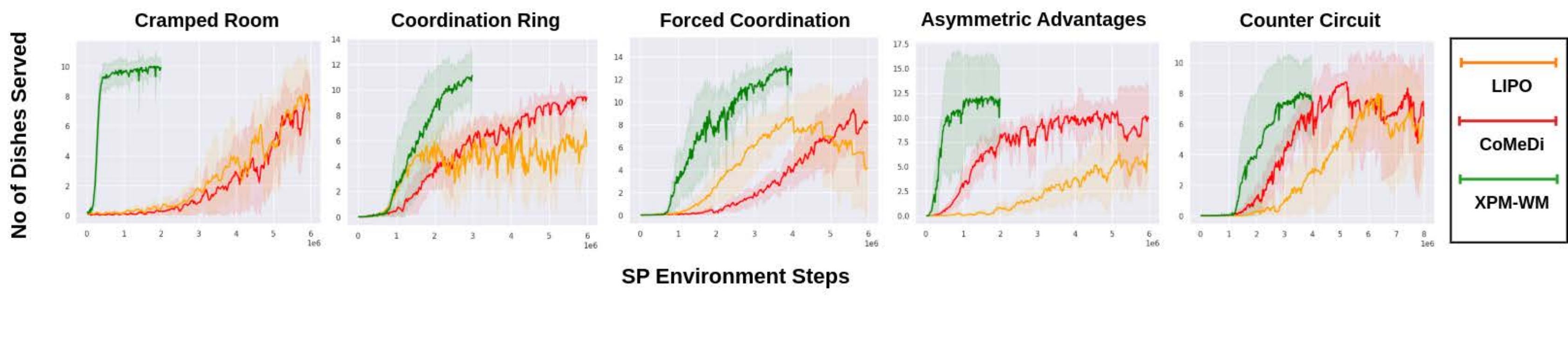}

% \vspace{-10mm}

\caption{Training Curves for all 5 Overcooked layouts. We plot the number of dishes served in each environment in each layout against the number of \textbf{SP training steps} (steps from XP and MP trajectories are excluded) averages across each subsequent agent in the population. The bold lines are the average dishes served across all subsequent agents and shaded area indicate the standard deviation.}

% \vspace{-3mm}
\label{fig:sp_training_curve}
\end{figure}

% \subsection{Pair-wise Cross-Play}

% \begin{figure*}[ht]
%   \centering
%   \subfigure[Cramped Room]{\includegraphics[width=0.35\linewidth]{images/cr_heatmap.pdf}}
%   \subfigure[Asymmetric Advantages]{\includegraphics[width=0.35\linewidth]{images/aa_heatmap.pdf}}
%   \subfigure[Coordination Ring]{\includegraphics[width=0.35\linewidth]{images/ring_heatmap.pdf}}
%   \subfigure[Forced Coordination]{\includegraphics[width=0.35\linewidth]{images/fc_heatmap.pdf}}
%   \subfigure[Counter Circuit]{\includegraphics[width=0.35\linewidth]{images/cc_heatmap.pdf}}
%   \caption{}.
%   \label{fig:spheatmap}
% \end{figure*}

% \clearpage

\subsection{Ego Agent Results for Specific Overcooked Layouts}
\label{ego_agent_res}
\begin{table*}[h]
    \centering
    \caption{Ego Agent Results for Cramped Room}
    \begin{tabular}{c|ccc}
        \toprule
        &E-LIPO &E-CoMeDi &E-XPM-WM \\
        \midrule
        LIPO     & $6.49 \pm 0.10$& $ 7.15\pm 0.10$&  $5.21 \pm 0.10$\\   
        CoMeDi  &   $8.20 \pm 0.10$& $7.44 \pm 0.12$&   $7.38 \pm 0.09$\\
        XPM-WM&  $6.35 \pm 0.11$& $6.49 \pm 3.07$& $8.52 \pm 0.08$\\
        \midrule
        Average &  $7.01 \pm 0.06$& $7.03 \pm 0.06$& $\bm{7.04 \pm 0.06}$\\
        \bottomrule
    \end{tabular}
    \label{tab:oc_ego_cr}
\end{table*}

\begin{table}[h]
    \centering
    \caption{Ego Agent Results for Asymmetric Advantages}
    \begin{tabular}{c|ccc}
        \toprule
        &E-LIPO &E-CoMeDi &E-XPM-WM \\
        \midrule
        LIPO     & $11.98 \pm 0.15$& $8.05 \pm 0.17$&  $8.34 \pm 0.17$\\   
        CoMeDi  &   $12.61 \pm 0.12$& $10.11 \pm 0.19$&   $9.59 \pm 0.16$\\
        XPM-WM&  $10.62 \pm 0.18$& $6.47 \pm 0.20$& $10.74 \pm 0.24$\\
        \midrule
        Average &  $\bm{11.74 \pm 0.09}$& $8.21 \pm  0.11$& $9.56 \pm  0.11$\\
        \bottomrule
    \end{tabular}
    % \caption{Asymmetric Advantages}
    \label{tab:oc_ego_aa}
\end{table}

\begin{table}[h]
    \centering
    \caption{Ego Agent Results for Coordination Ring}
    \begin{tabular}{c|ccc}
        \toprule
        &E-LIPO &E-CoMeDi &E-XPM-WM \\
        \midrule
        LIPO     & $3.99 \pm 0.07$& $3.02 \pm 0.09$&  $3.01 \pm 0.07$\\   
        CoMeDi  &   $5.32 \pm 0.07$& $4.91 \pm 0.10$&   $4.41 \pm 0.07$\\
        XPM-WM&  $4.14 \pm 0.09$& $1.12 \pm 0.06$& $4.69 \pm 0.09$\\
        \midrule
        Average &  $\bm{4.48 \pm  0.04}$& $3.02 \pm 0.06$& $4.04 \pm 0.05$\\
        \bottomrule
    \end{tabular}
    % \caption{Coordination Ring}
    \label{tab:oc_ego_cring}
\end{table}

\begin{table}[h]
    \centering
    \caption{Ego Agent Results for Forced Coordination}
    \begin{tabular}{c|ccc}
        \toprule
        &E-LIPO &E-CoMeDi &E-XPM-WM \\
        \midrule
        LIPO     & $3.75 \pm 0.12$& $6.26 \pm 0.11$&  $2.87 \pm 0.08$\\   
        CoMeDi  &   $1.58 \pm 0.08$& $5.20 \pm 0.09$&   $3.70 \pm 0.06$\\
        XPM-WM&  $1.30 \pm 0.06$& $3.56 \pm 0.06$& $4.81 \pm 0.12$\\
        \midrule
        Average &  $2.21 \pm 0.06$& $\bm{5.01 \pm 0.06}$& $3.70 \pm 0.06$\\
        \bottomrule
    \end{tabular}
    % \caption{Forced Coordination}
    \label{tab:oc_ego_fc}
\end{table}

\begin{table}[h]
    \centering
    \caption{Ego Agent Results for Counter Circuit}
    \begin{tabular}{c|ccc}
        \toprule
        &E-LIPO &E-CoMeDi &E-XPM-WM \\
        \midrule
        LIPO     & $2.09 \pm 0.07$& $2.31 \pm 0.06$&  $4.36 \pm 0.08$\\   
        CoMeDi  &   $4.41 \pm 0.10$& $3.60 \pm 0.11$&   $6.08 \pm 0.07$\\
        XPM-WM  &  $2.01 \pm 0.08$& $1.09 \pm 0.07$& $4.36 \pm 0.08$\\
        \midrule
        Average &  $2.75 \pm 2.61$& $2.33 \pm 2.72$& $\bm{4.66 \pm 2.56}$\\
        \bottomrule
    \end{tabular}
    % \caption{Counter Circuit}
    \label{tab:oc_ego_cc}
\end{table}

\clearpage

\section{Ablation Studies}

In this section we conduct additional ablation studies on different component of our proposed XPM-WM method. Specifically we show the effects of using vectorized reward prediction in the World Model. We then show the effects of using World Models on agent training, specifically the effects of the World Model as forward dynamics model of the environment. Finally we compare the wall clock time for XPM-WM as compared to other baseline methods 

\subsection{Effect of Vectorized Reward Prediction}

\begin{figure*}[ht!]
  \centering
  \fbox{\includegraphics[width=0.5 \linewidth]{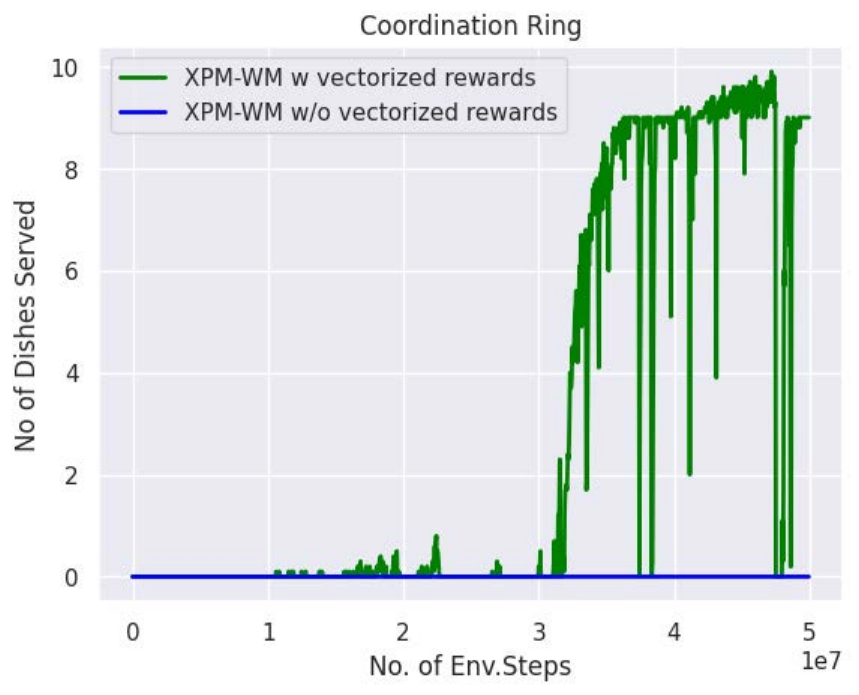}}
  \caption{A comparison of the training curves for XPM-WM with and without the use to vectorized event-based reward prediction on the Coordination Ring layout.}
  \label{fig:vec_rew_fig}
  % \vspace{-9pt}
\end{figure*}

Here we test the effect of using vectorized reward prediction when training the World Model. We include a variant of XPM-WM that predicts only a scalar joint reward at the reward predictor (modeled as Gaussian distribution) and compare with XPM-WM with vectorized rewards. We plot the training curves for Phase 1 of the training process in Figure \ref{fig:vec_rew_fig}. Our results show without explicitly factoring rewards, the world model fails to learn any meaningful policies.

\subsection{Effect of World Model as a Dynamics Model}

In this section we investigate the role of the World Model as a dynamics model on training efficiency.To do that we would need to decouple the dynamics modeling effects of the World Model from its role as a shared encoder for learning compact representations of the joint state. Hence we revisit MPPMR environment for this study. Specifically, we train a shared encoder network that is shared across all agents in the population while simulated trajectories can still be generated by the deterministic dynamics model that are separate from the agents. For the shared encoder, we first train the initial agent together with a Variational Autoencoder (VAE) \cite{kingma2013auto}. The VAE is trained to reconstruct the joint observation. The actor and critic takes in the latent representation of the VAE. The shared encoder is then used to bootstrap training for subsequent agents and VAE is still continuously updated similar to XPM-WM.

\begin{figure*}[ht!]
  \centering
  \fbox{\includegraphics[width=0.5 \linewidth]{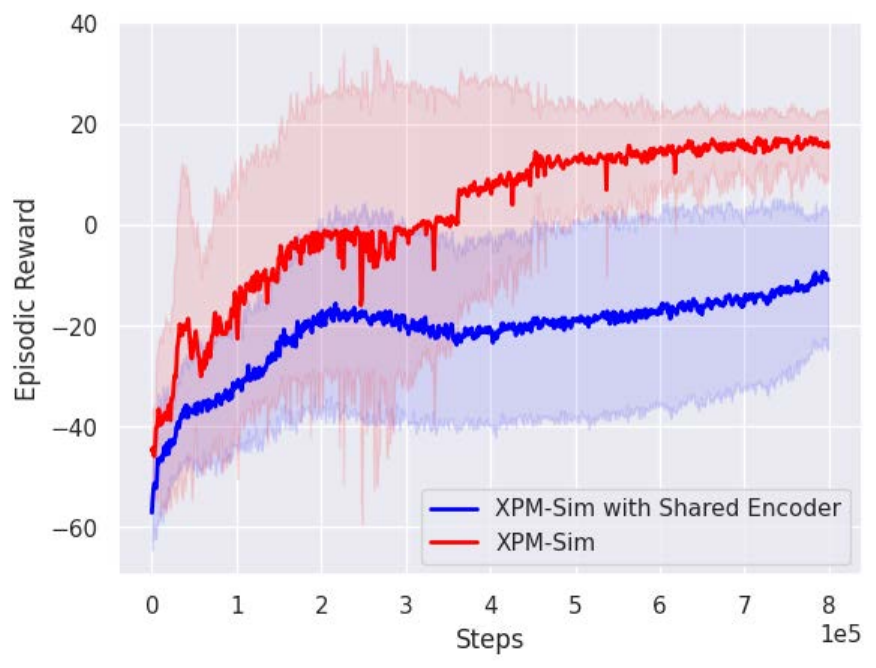}}
  \caption{A comparison of the training curves for agents 2-8 for XPM-Sim with and without the shared encoder on the MPPMR environment. For XPM-Sim with shared encoder, we first train the first agent together with a VAE. The VAE is trained to reconstruct the joint observation. The actor and critic takes in the latent representation of the VAE. The shared encoder is then used to bootstrap training for subsequent agents and VAE is still continuously updated similar to XPM-WM.}
  \label{fig:shared_encoder}
  % \vspace{-9pt}
\end{figure*}

We compare the training efficiency of this agent population to the one we describe in Section \ref{toy_exp} (separate agents without the shared encoder). We plot the average learning curves across all 8 agents in Figure \ref{fig:shared_encoder}. Our results show that agents trained with the shared encoder actually has a performance compared to agents trained without a shared encoder. We hypothesize is that this might be due to the fact that the shared encoder might have over-fitted to the state distribution of a particular (set of) partners and are therefore ill-suited as encoder for other partners which encounters states that are out of distribution for the encoder and it requires additional finetuning steps to re-align the encoder to the distributional shift, whereas a world model that is train to model the forward dynamics of the environment can still meaningfully represent states that are out if distribution.

\subsection{Training Wall Clock Time}
\label{wall_clock_time}

\begin{figure*}[ht!]
  \centering
  \fbox{\includegraphics[width=0.5 \linewidth]{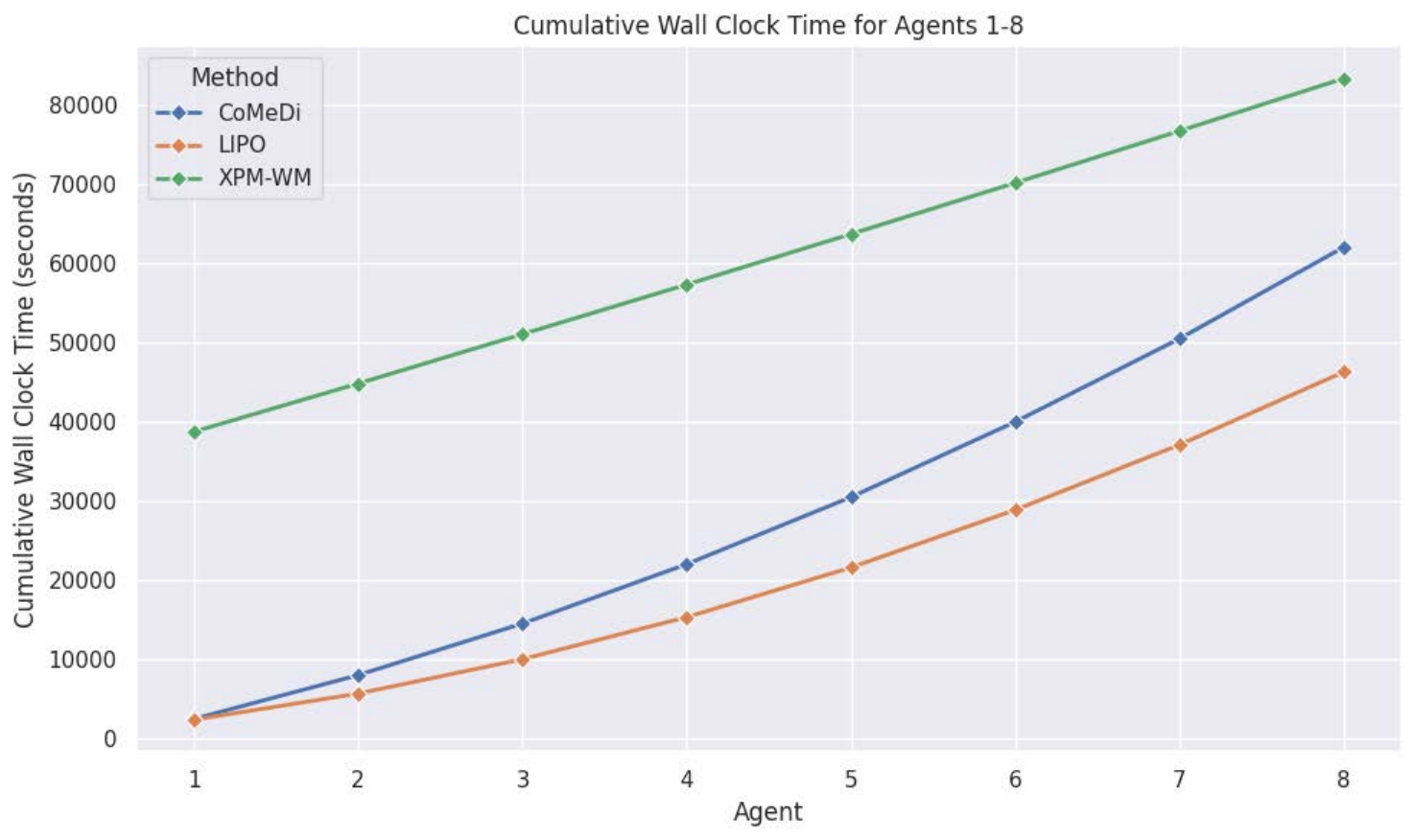}}
  \caption{A comparison of the training wall clock time for XPM-WM with CoMeDi and LIPO for all 8 agents on the Coordination Ring layout.}
  \label{fig:wall_clock_full}
  % \vspace{-9pt}
\end{figure*}

\begin{figure*}[ht!]
  \centering
  \fbox{\includegraphics[width=0.5 \linewidth]{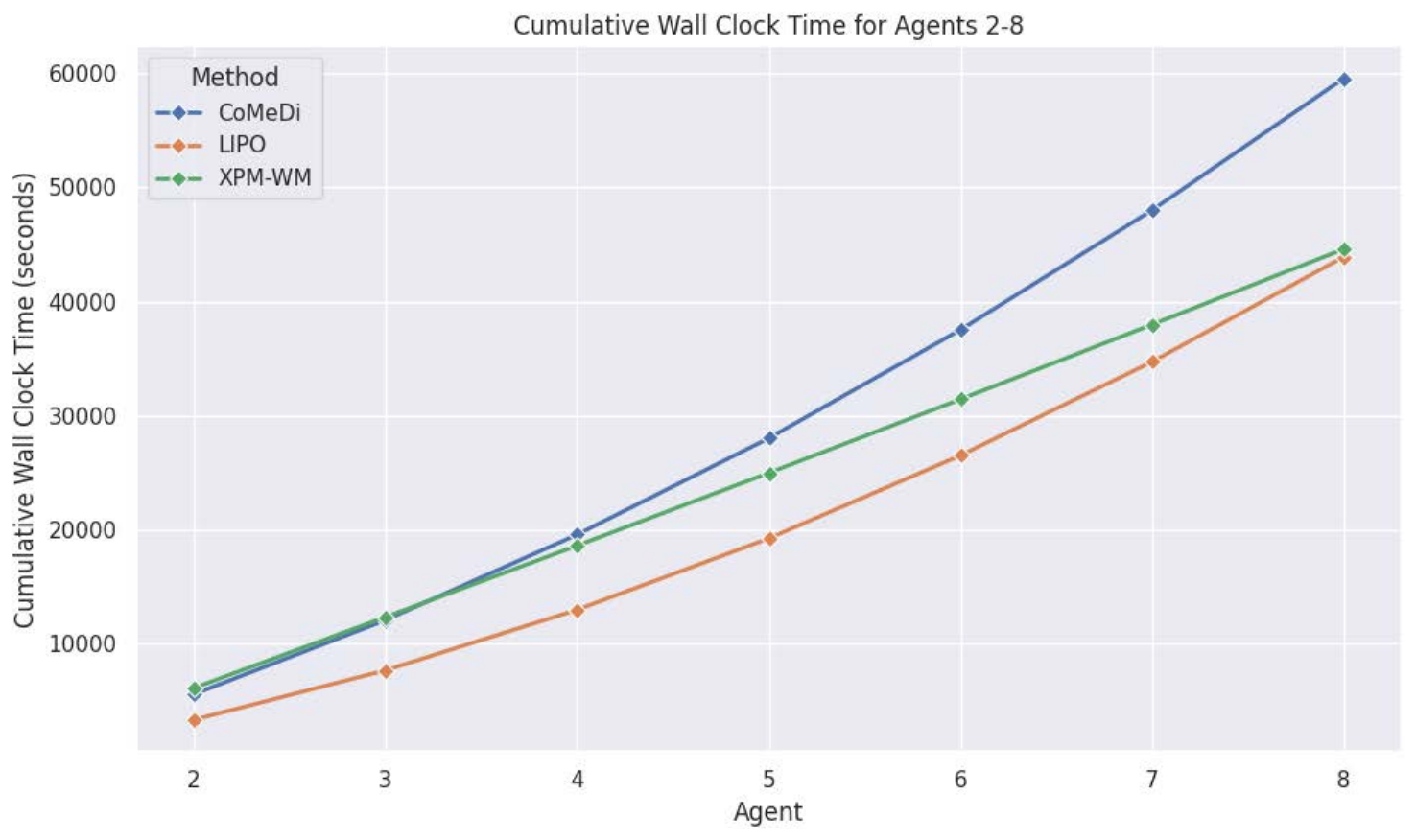}}
  \caption{A comparison of the training wall clock time for XPM-WM with CoMeDi and LIPO for agents 2-8 (Phase 2 training only) on the Coordination Ring layout.}
  \label{fig:wall_clock}
  % \vspace{-9pt}
\end{figure*}

In this section we compare the wall clock training time of XPM-WM as compared to CoMeDi and LIPO. We train all 8 agents on a single NVIDIA 4090 GPU and plot the cumulative wall clock time of training each subsequent agent. In Figure \ref{fig:wall_clock_full} we plot the wall clock time agents 1 to 8 which includes the World Model training for the initial agent. In Figure \ref{fig:wall_clock} we plot wall clock time for agents 2 to 8. We show that similar although training the World Model requires significantly more time compared to other baselines, subsequent agents actually are more time efficient as time cost for XPM-WM agents increase at a linear rate while agents trained using CoMeDi and LIPO increase at a steeper non-linear rate, suggesting the XPM-WM will overall also be more time efficient as we scale up the number of agents.

\end{document}